\documentclass[preprint,12pt]{elsarticle}
\usepackage{times}
\usepackage{amsmath}
\usepackage{amssymb}
\usepackage{graphicx}
\usepackage{xcolor}
\usepackage{mathrsfs}
\usepackage{caption}

\usepackage[bottom]{footmisc}
\usepackage{bm}
\usepackage{algorithm}
\usepackage{algpseudocode}

\usepackage{mathtools}
\usepackage{subcaption}

\usepackage{float}
\usepackage{adjustbox}
\usepackage[margin=1in]{geometry}
\usepackage[small,compact]{titlesec}
\usepackage[labelfont=bf,labelsep=colon,skip=2pt]{caption} 
\usepackage[titles]{tocloft}
\usepackage{bold-extra}
\usepackage{hyperref}

\hypersetup{
    colorlinks,
    linkcolor={red!50!black},
    citecolor={blue!50!black},
    urlcolor={blue!80!black}
}
\usepackage[left]{lineno}

\DeclareMathOperator*{\argmax}{arg\,max}
\DeclareMathOperator*{\tr}{tr}

\begin{document}
\begin{frontmatter}
\title{Bayesian Experimental Design for Model Discrepancy Calibration: An Auto-Differentiable Ensemble Kalman Inversion Approach}

\author{Huchen Yang}
\author{Xinghao Dong}
\author{Jin-Long Wu\corref{cor1}} \ead{jinlong.wu@wisc.edu} 
\cortext[cor1]{Corresponding author}

\address{Department of Mechanical Engineering, University of Wisconsin–Madison, Madison, WI 53706}

\begin{abstract}
Bayesian experimental design (BED) offers a principled framework for optimizing data acquisition by leveraging probabilistic inference. However, practical implementations of BED are often compromised by model discrepancy, i.e., the mismatch between predictive models and true physical systems, which can potentially lead to biased parameter estimates. While data-driven approaches have been recently explored to characterize the model discrepancy, the resulting high-dimensional parameter space poses severe challenges for both Bayesian updating and design optimization. In this work, we propose a hybrid BED framework enabled by auto-differentiable ensemble Kalman inversion (AD-EKI) that addresses these challenges by providing a computationally efficient, gradient-free alternative to estimate the information gain for high-dimensional network parameters. The AD-EKI allows a differentiable evaluation of the utility function in BED and thus facilitates the use of standard gradient-based methods for design optimization. In the proposed hybrid framework, we iteratively optimize experimental designs, decoupling the inference of low-dimensional physical parameters handled by standard BED methods, from the high-dimensional model discrepancy handled by AD-EKI. With the optimal designs identified for the model discrepancy, we can systematically obtain informative data to calibrate the model discrepancy. The performance of the proposed method is studied by a classical numerical example of BED governed by a convection-diffusion equation, and the results confirm that the hybrid BED framework enabled by AD-EKI efficiently identifies informative data to calibrate the model discrepancy and robustly infers the unknown physical parameters in the modeled system. Besides addressing the challenging problem of Bayesian experimental design with model discrepancy, AD-EKI can also potentially benefit the development of efficient and scalable frameworks in many other areas with bilevel optimization, such as meta-learning and structure optimization.
\end{abstract}

\begin{keyword}
Bayesian experimental design, Model discrepancy, Ensemble Kalman inversion, Differentiable programming, Bilevel optimization
\end{keyword}

\end{frontmatter}

\section{Introduction}

Bayesian experimental design (BED) provides a principled framework for optimizing data acquisition by leveraging Bayesian inference to guide the design process \cite{chaloner_bayesian_1995, lindley_measure_1956, lindley_bayesian_1972}. Rooted in information theory, BED aims to identify experimental conditions that maximize the information gain about target latent variables, typically by minimizing the expected posterior entropy or maximizing the expected Kullback–Leibler divergence between the prior and posterior distributions \cite{raiffa_applied_2000, degroot_optimal_2005, sebastiani_maximum_2000}. This probabilistic approach not only ensures that experimental designs are tailored to the specific goals of the study but also allows for adaptive, sequential experimentation, where newly acquired data refine subsequent designs \cite{ryan_review_2016, jones_bayes_2016, rainforth_modern_2023, huan_optimal_2024}. The flexibility and efficiency of BED have led to its successful application across diverse fields, including psychology \cite{myung_tutorial_2013, westfall_statistical_2014}, bioinformatics \cite{flynn_bayesian_2010, thompson_integrating_2023}, active learning \cite{golovin_near-optimal_2010, yang_active_2025}, medical imaging~\cite{go2025sequential}, and engineering systems~\cite{papadimitriou_optimal_2004, arroyave_perspective_2022,wu2023fast,chu2025bayesian}.

However, the practical implementation of BED is often hindered by model discrepancy, a phenomenon arising from the mismatch between the predictive model and the true physical system \cite{kennedy_bayesian_2001}. This discrepancy introduces systematic errors in the likelihood calculation, undermining the reliability of the posterior distribution and leading to suboptimal experimental designs. Specifically, model discrepancy can result in: (i) biased parameter estimates, as the inference process is based on an incorrect likelihood; and (ii) repeated selection of similar designs, yielding datasets that lack diversity and fail to capture the true system dynamics, ultimately diminishing the effectiveness of the entire experimental design process \cite{grunwald_inconsistency_2017, catanach_metrics_2023}. One prominent approach involves introducing correction terms and corresponding error parameters to approximate model discrepancy, thereby improving the fidelity of the likelihood function used in BED \cite{feng_optimal_2015}.

Recent progresses in data-driven modeling techniques have significantly advanced the calibration of model discrepancy by leveraging some empirically chosen data to refine model predictions \cite{kutz2016dynamic,peherstorfer2016data,wang2017physics,wu2018physics,duraisamy2019turbulence,montans_data-driven_2019,raissi2019physics,brunton2020machine,li2020fourier,lu2021learning,kochkov2021machine, levine2022framework, chen2025neural}. Although a BED framework has the potential of systematically identifying informative data for the calibration of some unknown parameters, the model discrepancy is often of a strongly nonlinear form and characterized by high-dimensional unknown parameters~\cite{chen2024cgnsde, dong2024data}, which poses a key challenge to standard Bayesian inference methods. While recent work has demonstrated that designs and data optimized for low-dimensional physical parameters of interest can also inform the calibration of the model discrepancy, additional evaluations with an empirically chosen threshold are often required to exclude misleading data and ensure robust performance \cite{yang_active_2025}. On the other hand, classical Bayesian inference methods for information gain estimation consist of Monte Carlo-based approaches \cite{metropolis_equation_1953, gelfand_sampling-based_1990, hastings_monte_1970}, such as Rao-Blackwellized Monte Carlo methods \cite{gal_deep_2017, foster_variational_2021} and nested Monte Carlo estimators \cite{rainforth_nesting_2018}, which are computationally expensive for high-dimensional problems and thus motivate the development of alternative and more affordable approaches, such as variational approaches  \cite{dong_variational_2025, foster_variational_2019} and Laplace approximation \cite{laplace_memoir_1986, tierney_accurate_1986}.

Monte Carlo (MC) methods, despite their theoretical robustness, suffer from high computational costs due to the need for nested sampling in expected information gain (EIG) estimation, leading to an exponential increase in complexity \cite{wakefield_expected_1994, palmer_bayesian_1998, robert_monte_2004}. The significant computational cost of MC methods also imposes a challenge to the optimization methods for search good designs. Although recent stochastic gradient descent (SGD) has improved computational efficiency by estimating EIG gradients through a small number of samples \cite{huan_gradient-based_2014, foster_unified_2020, carlon_nesterov-aided_2020}, an accurate evaluation of utility function for high-dimensional unknown parameters still demands a huge amount of samples, while the approximation via a small number of random samples could still lead to non-negligible biases. In addition, methods used to overcome the instability of gradient-based optimization for optimal design under the MC framework typically suffer from high variance, increased per-sample computational costs, or scalability issues in high-dimensional settings \cite{giles_multilevel_2008, snyder_obstacles_2008, goda_unbiased_2022}. On the contrary, gradient-free approaches, such as Bayesian optimization \cite{foster_variational_2019, kleinegesse_efficient_2019} and evolutionary algorithms \cite{hamada_finding_2001, price_induced_2018}, can be employed for design selection but generally require an excessive number of function evaluations, making them impractical for complex and large models in many real-world applications.

Variational inference (VI) provides a computationally efficient alternative to MC approaches by approximating the posterior distribution through optimization \cite{blei_variational_2017, ormerod_explaining_2010}. Such methods typically exploit reparameterization tricks that facilitate low-variance, gradient-based optimization via automatic differentiation \cite{rezende_stochastic_nodate, chen_auto-encoding_2024}. However, their effectiveness critically depends on the forward model being differentiable with respect to the parameters of interest \cite{krishnan_structured_2017, poole_variational_2019, chen_autodifferentiable_2022}, while gradient-free optimization methods tend to exhibit high variance, leading to slower convergence and decreased stability, particularly in high-dimensional scenarios \cite{williams_simple_1992, ranganath_black_2013, tucker_rebar_2017}. More importantly, since VI is formulated as an optimization problem, extending the framework to further optimize design variables results in a nested optimization structure. In such cases, the inner VI optimization can hinder gradient propagation to the outer level optimization, while also introducing additional computational burden and numerical instability. To enhance computational efficiency in the nested design optimization, recent approaches have focused on frameworks that bypass direct differentiation with respect to design variables by jointly optimizing the posterior approximation and design variables through maximizing a lower bound on the information gain, thereby avoiding direct divergence computations and achieving more stable gradient estimates \cite{foster_unified_2020}. However, like all variational inference methods, the effectiveness of these strategies depends on the quality of the variational approximation, necessitating careful selection of the approximating family \cite{sobolev_importance_2019, ambrogioni_automatic_2021}. 

Laplace approximation also serves as a computationally efficient alternative to MC methods and has been explored in the Bayesian optimal design. The core idea of the Laplace approximation is to approximate the posterior distribution by a Gaussian distribution centered around the maximum a posteriori (MAP) estimate \cite{tierney_accurate_1986}. This Gaussian assumption allows one to replace the otherwise intractable integration with a local second-order Taylor expansion in closed-form, which transforms the inner-loop Bayesian update, originally formulated as an integration problem, into an optimization problem \cite{dodds_robust_2005}. This change can significantly reduce the computational cost of posterior estimation, thereby lowering the per-iteration expense of the outer-loop design optimization \cite{beck_fast_2018, ryan_fully_2015}. However, the effectiveness of the Laplace approximation relies on the local Gaussianity assumption, which may break down for multimodal or highly skewed posteriors. Recent developments in Integrated Nested Laplace Approximation (INLA,~\cite{rue_approximate_2009, long_fast_2013}) have been shown to mitigate these issues by providing more accurate posterior estimates, even in complex models. However, by transforming the inner posterior update into an optimization problem, these methods inherently introduce a nested (or bilevel) optimization structure, similar to variational approaches. This nested structure poses additional challenges for the optimization algorithms, as the outer-loop design optimization must now account for the computational cost and potential convergence issues of the inner-loop optimization.

Instead of employing variational inference or Laplace approximation, we aim to provide a new perspective for BED, by exploring ensemble Kalman inversion (EKI, \cite{iglesias_ensemble_2013}) to build a framework that efficiently identifies optimal designs and data for the calibration of model discrepancy, which is often parametrized by high-dimensional unknown parameters. Unlike the VI or Laplace approximation methods that rely on the gradient descent optimization, ensemble Kalman inversion is derivative-free, parallelizable, and robust with noises of the data and chaos or uncertainties of the system~\cite{schneider2021learning,wu2024learning}. In addition, empirical results~\cite{kovachki_ensemble_2019} also demonstrated that EKI is efficient and robust to handle high-dimensional and strongly nonlinear problems, despite its Gaussian assumption. In the past decade, developments have been made to enhance Kalman inversion methods in establishing the theoretical foundation~\cite{schillings2017analysis,ding2021ensemble,calvello2022ensemble}. In the context of experimental design, it is worth noting that if the inner-loop Bayesian update is fully differentiable, then the outer-loop design optimization can leverage efficient gradient-based methods without being hindered by the challenges typically associated with nested optimization. Recent work on auto-differentiable Ensemble Kalman Filters (AD-EnKF,~\cite{chen_autodifferentiable_2022}) pioneered the development of auto-differentiable ensemble Kalman methods and inspired us to explore an auto-differentiable EKI for the inner optimization problem of BED. In fact, for any other bilevel optimization problems, employing EKI in the inner optimization loop can facilitate adopting existing tools of differentiable programming for the outer-loop optimization and potentially offer computational benefits and more robust optimization performance.

In this work, we propose a hybrid BED framework enabled by auto-differentiable ensemble Kalman inversion (AD-EKI), to address the challenge of classical BED approaches associated with Bayesian inference for high-dimensional network parameters. This approach is featured by the efficient and differentiable approximation of utility function in BED, thereby enabling the use of standard gradient-based techniques for the optimization of designs. More specifically, the proposed hybrid BED framework assumes two types of unknown parameters, including low-dimensional parameters of the physics-based model and high-dimensional parameters that parametrize the model discrepancy. We apply the framework to iteratively estimate the physical parameters and the model discrepancy as shown in Fig.~\ref{fig:graphic_abstract}. For the low-dimensional physical parameters, we employ a traditional BED approach to determine optimal designs and the corresponding data, followed by a full Bayesian update to refine their belief based on the current model. For high-dimensional parameters of the model discrepancy, we employ AD-EKI to efficiently approximate the utility function that quantifies the information gain, based on the maximum a posteriori (MAP) estimate of the physical parameters; this utility is then maximized to guide the next experimental design and drive a deterministic update of the discrepancy network parameters.
 
The key highlights of this work are as follows.
\begin{itemize}
    \item We introduce an ensemble-Kalman-based approach that leverages a Gaussian assumption on high-dimensional parameters of model discrepancy to provide an efficient alternative to full Bayesian inference in the evaluation of information gain for Bayesian experimental design.
    \item We develop an auto-differentiable framework that integrates the inherently gradient-free ensemble Kalman inversion, enabling the use of standard gradient-based optimization techniques of experimental designs for high-dimensional parameters that parametrize the model discrepancy.
    \item We study the computational cost and the scalability of the auto-differentiable ensemble Kalman inversion in terms of ensemble size and total iterations, which can potentially benefit other challenging bilevel optimization problems.
\end{itemize}

This paper is organized as follows. Section \ref{sec: Methodology} introduces the general formulation of our method. Section \ref{sec: Numerical Results} provides numerical results of a classical example governed by a convection-diffusion equation and used for testing the performance of BED methods. Finally, Section \ref{Conclusion} concludes the paper.

\section{Methodology}\label{sec: Methodology}

The difference between the true system and a physics-based model, known as model discrepancy, is ubiquitous in the modeling and simulation of complex dynamical systems~\cite{kennedy_bayesian_2001, brynjarsdottir2014learning}. Various data-driven methods have been explored to learn model discrepancy, improving state forecasts or parameter estimates~\cite{luo2020accounting, dong2025stochastic}. Existing data-driven methods on the calibration of model discrepancy often rely on empirical choices of the training data rather than systematically designing it. In contrast to the physical parameters of a model, which typically reside in a low-dimensional vector space, the model discrepancy is an element of an infinite-dimensional function space~\cite{wu2024learning}. Consequently, approximating this function with methods such as neural networks leads to a high-dimensional coefficient space, which poses a significant challenge for systematically identifying optimal training data using standard BED methods~\cite{yang_active_2025}. In this work, we aim to build on the AD-EKI and develop an efficient framework to identify optimal designs that guide active data acquisition for the calibration of the model discrepancy.

The true system in a general form can be written as:
\begin{equation}
\label{eq:true_system}
\begin{aligned}
    \frac{\partial \mathbf{u}}{\partial t} = \mathcal{G}^\dagger(\mathbf{u}; \boldsymbol{\theta}^\dagger),
\end{aligned}
\end{equation}
where the system state $\mathbf{u}(\mathbf{z}, t)$ is a spatiotemporal field and $\mathcal{G}^\dagger$ denotes the true dynamics. Accurate estimation of the unknown parameters $\boldsymbol{\theta}^\dagger$ could be achieved by BED only when the detailed form of true dynamics is known. In practice, the form of true dynamics is often unknown for many complex dynamical systems. Assuming that a physics-based model $\mathcal{G}(\mathbf{u};\boldsymbol{\theta}_\mathcal{G})$ exists to approximate the true dynamics $\mathcal{G}^\dagger$, we focus on the learning of model discrepancy $\mathcal{G}^\dagger - \mathcal{G}$, which is characterized by a neural network $\textrm{NN}(\mathbf{u};\boldsymbol{\theta}_\textrm{NN})$ and leads to the modeled system:

\begin{equation}
\label{eq:modeled_system}
    \frac{\partial \mathbf{u}}{\partial t}=\mathcal{G}(\mathbf{u};\boldsymbol{\theta}_\mathcal{G})+\textrm{NN}(\mathbf{u};\boldsymbol{\theta}_\textrm{NN}).
\end{equation}

The neural network $\text{NN}$ is trained using measurement data $\mathbf{y}$, which can be viewed as a function of the experimental design $\mathbf{d}$ in the context of BED. In this work, the design $\mathbf{d}$ corresponds to the spatiotemporal coordinate $(\mathbf{z}, t)$. The measurement $\mathbf{y}$ is the observation of variable $\mathbf{u}$ that represents the state of the physical system at the design $\mathbf{d}$, i.e., $\mathbf{y} = \mathbf{u}(\mathbf{d}) + \boldsymbol\epsilon$, where $\boldsymbol\epsilon \sim \mathcal{N}(\boldsymbol{0}, \boldsymbol{\Sigma}_{\boldsymbol{\epsilon}})$ represents the measurement noise which is assumed to be Gaussian in this work when observing the state from the true system.

\subsection{Bayesian Experimental Design}
\label{sec:OED}

The Bayesian experimental design~\cite{lindley_bayesian_1972, rainforth_modern_2023,huan_optimal_2024} provides a general framework to systematically seek the optimal design by solving the optimization problem: 
\begin{equation}
\label{eq:oed_opt}
    \begin{aligned}
        \mathbf{d}^* &= \argmax_{\mathbf{d}\in\mathcal{D}} \mathbb{E}[U(\boldsymbol{\theta},\mathbf{y},\mathbf{d})]\\
        &= \argmax_{\mathbf{d}\in\mathcal{D}} \int_\mathcal{Y} \int_{\boldsymbol{\Theta}} U(\boldsymbol{\theta},\mathbf{y},\mathbf{d}) p(\boldsymbol{\theta},\mathbf{y}|\mathbf{d}) \mathrm{d}\boldsymbol{\theta}\mathrm{d}\mathbf{y},
    \end{aligned}
\end{equation}
where $\mathbf{y} \in \mathcal{Y} \subset \mathbb{R}^{d_{\mathbf{y}}}$ is data from the experimental design $\mathbf{d} \in \mathcal{D}\subset \mathbb{R}^{d_{\mathbf{d}}}$, $\boldsymbol{\theta} \in \boldsymbol{\Theta}\subset \mathbb{R}^{d_{\boldsymbol{\theta}}}$ denotes the target parameters, and $U$ is the utility function taking the inputs of $\mathbf{y}$ and $\boldsymbol{\theta}$ and returning a real value, which reflects the specific purpose of designing the experiment. The optimal design $\mathbf{d}^*$ is obtained by maximizing the expected utility function $\mathbb{E}[U(\boldsymbol{\theta},\mathbf{y},\mathbf{d})]$ over the design space $\mathcal{D}$. The term $p(\boldsymbol{\theta},\mathbf{y}|\mathbf{d})$ is the joint conditional distribution of data and parameters.

The utility function $U(\boldsymbol{\theta},\mathbf{y},\mathbf{d})$ can be regarded as the information gain obtained from the data $\mathbf{y}$ for the corresponding design $\mathbf{d}$. This work employs Kullback-Leibler (KL) divergence between the posterior and the prior distribution of parameters $\boldsymbol{\theta}$:
\begin{equation}
    \begin{aligned}
        U(\boldsymbol{\theta},\mathbf{y},\mathbf{d}) &= D_{\textrm{KL}} \bigl(p(\boldsymbol\theta|\mathbf{y},\mathbf{d})\| p(\boldsymbol\theta)\bigr)\\
        &=\mathbb{E}_{\boldsymbol\theta|\mathbf{d},\mathbf{y}}(\log p(\boldsymbol{\theta}|\mathbf{y},\mathbf{d}) - \log p(\boldsymbol{\theta}))\\
        &=\int_{\boldsymbol{\Theta}} p(\boldsymbol{\theta}|\mathbf{y},\mathbf{d}) \log(\frac{p(\boldsymbol{\theta}|\mathbf{y},\mathbf{d})}{p(\boldsymbol{\theta})})\mathrm{d}\boldsymbol{\theta},
    \end{aligned}
    \label{eq: kld utility}
\end{equation}
where $p(\boldsymbol\theta|\mathbf{y},\mathbf{d})$ is the posterior distribution of $\boldsymbol\theta$ given a design $\mathbf{d}$ and the data $\mathbf{y}$. Note that data $\mathbf{y}$ could be either from the actual experimental measurement or the numerical model simulation. The use of KL divergence as the utility function leads to the definition of expected information gain (EIG,~\cite{lindley_measure_1956}) that integrates the information gain over all possible predicted data:
\begin{equation}
\label{eq:EIG}
    \begin{aligned}
        \text{EIG} (\mathbf{d}) &=\mathbb{E}_{\mathbf{y} | \mathbf{d}} [D_{\textrm{KL}}(p(\boldsymbol\theta|\mathbf{y},\mathbf{d})\| p(\boldsymbol\theta))]\\
  &=\int_{\mathcal{Y}}D_{\textrm{KL}}(p(\boldsymbol\theta|\mathbf{y},\mathbf{d})\| p(\boldsymbol\theta)) p(\mathbf{y}|\mathbf{d})\mathrm{d}\mathbf{y},\\      
    \end{aligned}
\end{equation}
where $p(\mathbf{y}|\mathbf{d}):=\mathbb{E}_{\boldsymbol{\theta}}[p(\mathbf{y}|\boldsymbol{\theta},\mathbf{d})]$ is the distribution of the predicted data among all possible parameter values given a certain design.

The performance of BED relies on the accuracy of forward model~\cite{rainforth_modern_2023, catanach_metrics_2023}, while the model discrepancy~\cite{kennedy_bayesian_2001,feng_optimal_2015} is inevitable in most real-world applications. Although the recent developments of residual learning~\cite{levine2022framework, yang_active_2025} are promising for characterizing the model discrepancy of BED, they often involve neural networks with a large number of unknown coefficients, for which systematically gathering informative training data via a full Bayesian perspective (e.g., MC methods) is often computationally infeasible. Unlike MC methods, variational methods provide more efficient, gradient-based alternatives for approximating the posterior distribution. However, when further optimizing the design based on information entropy in the outer-loop, the direct application of gradient-based variational methods in the inner-loop remains challenging, limiting their efficiency in practical BED implementations and potentially leading to an inconsistent formulation compared to the original bilevel optimization problem. On the other hand, Laplace approximation~\cite{tierney_accurate_1986,long_fast_2013,long_laplace_2015,beck_fast_2018,bartuska2025laplace} has been explored to approximately solve the Bayesian inference problem involved in BED, while the intrinsic Gaussian assumption could be too restrictive in addressing the neural-network-based model discrepancy that often implies strong nonlinearity. Therefore, we aim to propose an alternative approach to the bilevel optimization problem of BED for model discrepancy, by developing an ensemble Kalman method for the inner-loop optimization that can handle nonlinearity and supports automatic differentiation, to facilitate computing gradients of information gain with respect to the experimental design for the outer-loop optimization.

\subsection{Ensemble-based Utility Function}\label{Sec: ensemble-based utility}
The general concept of this work is to employ an ensemble Kalman method to efficiently approximate the posterior distribution and further calculate information gain, which accounts for the main computational challenge to standard BED methods when $\boldsymbol{\theta}$ is high-dimensional. More specifically, ensemble Kalman inversion (EKI) is employed to approximately quantify the information gain of a given data point for the calibration of $\boldsymbol{\theta}_\text{NN}$. EKI was originally developed to solve a Bayesian inverse problem in the general form:
\begin{equation}
    \label{eq:eki_inverse_problem}
    \mathbf{y}=G(\boldsymbol\theta,\mathbf{d})+\boldsymbol\eta,
\end{equation}
where $G$ denotes a forward map from unknown parameters $\boldsymbol{\theta}$ to the data $\mathbf{y}$, and $\boldsymbol\eta \sim \mathcal{N}(\mathbf{0},\Gamma)$ denotes the data measurement noises and is often assumed to be with a zero-mean Gaussian distribution. 

Given a design $\mathbf{d}$ and the corresponding data measurement $\mathbf{y}$, the EKI updating formula for the ensemble of parameters can be written as:
\begin{equation}
\boldsymbol\theta_{n+1}^{(j)}=\boldsymbol\theta_{n}^{(j)}+\Sigma^{\boldsymbol\theta \mathbf{g}}_n(\Sigma^{\mathbf{g}\mathbf{g}}_n+\Gamma)^{-1}(\mathbf{y}^{(j)}-\mathbf{g}^{(j)}_n),
\label{eq: eki updating}
\end{equation}
where $\mathbf{y}^{(j)}=\mathbf{y}+\epsilon,~\epsilon\sim \mathcal{N}(0,\Gamma)$ is the perturbed observation and $\mathbf{g}^{(j)}_n:=G(\boldsymbol\theta_{n}^{(j)},\mathbf{d})$, the index $n$ denotes the $n$-th EKI iteration, and the index $j$ indicates the $j$-th ensemble. The ensemble covariance matrices $\Sigma^{\boldsymbol\theta \mathbf{g}}_n$ and $\Sigma^{\mathbf{g}\mathbf{g}}_n$ can be calculated as:
\begin{equation}
    \begin{aligned}
    &\bar{\boldsymbol\theta}_n = \frac{1}{J}\sum_{j=1}^J \boldsymbol\theta_n^{(j)}, \quad \bar{\mathbf{g}}_n = \frac{1}{J}\sum_{j=1}^J \mathbf{g}^{(j)}_n,  \\
    &\Sigma^{\boldsymbol\theta \mathbf{g}}_n = \frac{1}{J-1} \sum_{j=1}^J \left(\bar{\boldsymbol\theta}_n - \boldsymbol\theta_n^{(j)}\right)\left(\bar{\mathbf{g}}_n  - \mathbf{g}^{(j)}_n \right)^\top ,   \\
    &\Sigma^{\mathbf{g}\mathbf{g}}_n = \frac{1}{J-1} \sum_{j=1}^J \left(\bar{\mathbf{g}}_n  - \mathbf{g}^{(j)}_n \right)\left( \bar{\mathbf{g}}_n  - \mathbf{g}^{(j)}_n \right)^\top.
    \label{eq: eki detials}
\end{aligned}
\end{equation}

With the initial ensemble $\{\boldsymbol\theta_0^{(j)}\}^J_{j=1}$ and the updated ensemble $\{\boldsymbol\theta_K^{(j)}\}^J_{j=1}$ after $K$ times of EKI iterations, we follow the standard information-based BED paradigm~\cite{chaloner_bayesian_1995,lindley_bayesian_1972} and quantify the information gain using the KL divergence between these two ensembles. By assuming each ensemble is drawn from a multivariate Gaussian distribution, this ensemble-based KL admits a closed-form approximation:
\begin{equation}
\begin{gathered}
\label{eq:ensemble_KLD}
        D_{\textrm{KL}}(p(\boldsymbol{\theta}|\mathbf{y}) || p(\boldsymbol{\theta})  ) \approx \tilde{D}_{\textrm{KL}}(\{\boldsymbol\theta_K^{(j)}\}^J_{j=1} || \{\boldsymbol\theta_0^{(j)}\}^J_{j=1}) = \\ 
        \frac{1}{2} \left[ \text{tr} \left( \left( \Sigma^{\boldsymbol{\theta} \boldsymbol{\theta}}_0 \right)^{-1} \Sigma^{\boldsymbol{\theta} \boldsymbol{\theta}}_K \right) - d_{\boldsymbol\theta} + \ln \left( \frac{\det \Sigma^{\boldsymbol{\theta} \boldsymbol{\theta}}_0}{\det \Sigma^{\boldsymbol{\theta} \boldsymbol{\theta}}_K} \right)\right. 
        \left.+ \left(\bar{\boldsymbol{\theta}}_0 - \bar{\boldsymbol{\theta}}_K \right)^\top \left( \Sigma^{\boldsymbol{\theta} \boldsymbol{\theta}}_0 \right)^{-1} \left(\bar{\boldsymbol{\theta}}_0 - \bar{\boldsymbol{\theta}}_K \right) \right].
\end{gathered}
\end{equation}

where $d_{\boldsymbol{\theta}}$ is the dimension of parameter $\boldsymbol{\theta}$, and the ensemble covariance matrices $\Sigma^{\boldsymbol\theta \boldsymbol\theta}_0$ and $\Sigma^{\boldsymbol\theta \boldsymbol\theta}_K$ can be calculated as:
\begin{align}
    &\Sigma^{\boldsymbol\theta \boldsymbol\theta}_n = \frac{1}{J-1} \sum_{j=1}^J \left(\bar{\boldsymbol\theta}_n - \boldsymbol\theta_n^{(j)}\right)\left(\bar{\boldsymbol\theta}_n - \boldsymbol\theta_n^{(j)}\right)^\top.
\end{align}

In summary, we employ EKI to update the ensemble of unknown parameters $\boldsymbol{\theta}$, transforming the initial ensemble $\{\boldsymbol\theta_0^{(j)}\}^J_{j=1}$ into the updated ones $\{\boldsymbol\theta_K^{(j)}\}^J_{j=1}$, informed by the observation data $\mathbf{y}$ and model prediction $\mathbf{g}^{(j)}_n$. The divergence between these two sets of ensembles provides an approximation $\tilde{D}_\text{KL}$ of the information gain $D_{\textrm{KL}}$ for the given observation data $\mathbf{y}$. Within the context of experimental design, the design variable $\mathbf{d}$ dictates the observations $\mathbf{y}(\mathbf{d})$ and the model predictions $\mathbf{g}^{(j)}_n(\boldsymbol\theta_{n}^{(j)},\mathbf{d})$, thereby establishing a mapping $\mathbf{d} \mapsto \tilde{D}_\text{KL}$. The ensemble-based approximation of KL divergence in Eq.~\eqref{eq:ensemble_KLD} provides an efficient means to quantify the informativeness of the data $\mathbf{y}$ from a design $\mathbf{d}$. This formulation of KL divergence can be directly integrated into the EIG, and its evaluation can be efficiently implemented with parallel computing.

\subsection{AD-EKI for A Differentiable Estimation of KL Divergence}
\label{sec:AD_EKI_KL}
EKI was originally developed to solve the Bayesian inverse problem in Eq.~\eqref{eq:eki_inverse_problem} via a derivative-free perspective, i.e., the gradient information is not required for the forward map $G$ with respect to the unknown parameters. By employing EKI to approximately solve the Bayesian inference problem involved in the definition of information gain for BED, it is worth noting that implementation of auto-differentiable EKI can facilitate the standard gradient-based optimization of BED to search for the optimal design, i.e., the optimization problem defined in Eq.~\eqref{eq:oed_opt}. Generally speaking, the EKI updating formulae summarized in Section~\ref{Sec: ensemble-based utility} only involve the ensemble evaluation of forward map $G$ and the matrix-vector manipulations, for which an auto-differentiable version can be implemented based on existing differentiable programming tools such as JAX.

In general, the ensemble-based KL divergence explicitly depends on the distribution of the ensemble at each EKI iteration, i.e., $\bar{\boldsymbol{\theta}}_n$ and $\Sigma_n^{\boldsymbol{\theta}\boldsymbol{\theta}}$, which further depends on the input data $\mathbf{y}(\mathbf{d})$ and the predictions $\mathbf{g}^{(j)}_{n-1}(\boldsymbol{\theta}^{(j)}_{n-1},\mathbf{d})$. Here we present a brief derivation for the detailed form of $\mathrm{d} \tilde{D}_\text{KL} / \mathrm{d} \mathbf{d}$ to facilitate the discussion of computational complexity. Note that all gradient calculations with respect to the design $\mathbf d$ are taken on the ensemble-based utility $\tilde{D}_{\mathrm{KL}}(\mathbf d)$. More specifically, considering the $K$-th iteration of EKI, the terms in Eq.~\eqref{eq:ensemble_KLD} that are dependent on $\mathbf{d}$ include the ensemble mean $\bar{\boldsymbol{\theta}}_K$ and the covariance matrix $\Sigma_K^{\boldsymbol{\theta}\boldsymbol{\theta}}$. The initial ensemble parameters, $\bar{\boldsymbol{\theta}}_0$ and $\Sigma_0^{\boldsymbol{\theta}\boldsymbol{\theta}}$, are pre-specified and independent of the design $\mathbf{d}$. The gradient $\nabla_\mathbf{d} \tilde{D}_\text{KL}(\mathbf{d})$ of the ensemble-based KL divergence with respect to the design is explicitly given by

\begin{equation}
    \begin{aligned}\label{eq: EKI gradient 1}
        \frac{\mathrm{d} \tilde{D}_\text{KL}}{\mathrm{d} \mathbf{d}}=&\frac{\partial \tilde{D}_\text{KL}}{\partial \bar{\boldsymbol{\theta}}_K} \frac{\mathrm{d} \bar{\boldsymbol{\theta}}_K}{\mathrm{d} \mathbf{d}} + \frac{\partial \tilde{D}_\text{KL}}{\partial \Sigma^{\boldsymbol{\theta}\boldsymbol{\theta}}_K} : \frac{\mathrm{d} \Sigma^{\boldsymbol{\theta}\boldsymbol{\theta}}_K}{\mathrm{d} \mathbf{d}}\\
        =&(\Sigma^{\boldsymbol{\theta}\boldsymbol{\theta}}_0)^{-1} (\bar{\boldsymbol{\theta}}_K - \bar{\boldsymbol{\theta}}_0)\cdot\frac{\mathrm{d} \bar{\boldsymbol{\theta}}_K}{\mathrm{d} \mathbf{d}} + \frac{1}{2}[ (\Sigma^{\boldsymbol{\theta}\boldsymbol{\theta}}_0)^{-1}- (\Sigma^{\boldsymbol{\theta}\boldsymbol{\theta}}_K)^{-1}] : \frac{\mathrm{d} \Sigma^{\boldsymbol{\theta}\boldsymbol{\theta}}_K}{\mathrm{d} \mathbf{d}},
    \end{aligned}
\end{equation}
where $:$ represents the Frobenius product. Detailed derivation of Eq.~\eqref{eq: EKI gradient 1} can be found in \ref{apd:derivation}. The gradient terms at the right-hand side of Eq.~\eqref{eq: EKI gradient 1} can be further expanded as
\begin{equation}
   \begin{aligned}\label{eq: EKI gradient 2}
       \frac{\mathrm{d} \bar{\boldsymbol{\theta}}_K}{\mathrm{d} \mathbf{d}}  &= \frac{\mathrm{d} }{\mathrm{d} \mathbf{d}}  \left[\bar{\boldsymbol{\theta}}_0 + \sum_{n=0}^{K-1} \mathbf{K}_n(\mathbf{d})(\mathbf{y}(\mathbf{d})-\bar{\mathbf{g}}_n(\mathbf{d}))\right] =\sum_{n=0}^{K-1} \left[\frac{\mathrm{d} \mathbf{K}_n}{\mathrm{d} \mathbf{d}} (\mathbf{y}-\bar{\mathbf{g}}_n) + \mathbf{K}_n \left(\frac{\mathrm{d} \mathbf{y}}{\mathrm{d} \mathbf{d}}-\frac{\mathrm{d} \bar{\mathbf{g}}_n}{\mathrm{d} \mathbf{d}}\right) \right],\\
       \frac{\mathrm{d} \Sigma_K^{\boldsymbol{\theta}\boldsymbol{\theta}}}{\mathrm{d} \mathbf{d}}&=  \frac{\mathrm{d} }{\mathrm{d} \mathbf{d}} \left[ \Sigma_0 - \sum_{n=0}^{K-1}\mathbf{K}_n(\mathbf{d})\Sigma_n^{\boldsymbol{\theta}\mathbf{g}}(\mathbf{d})\right]= \sum_{n=0}^{K-1} \left[ - \frac{\mathrm{d} \mathbf{K}_n}{\mathrm{d} \mathbf{d}} \Sigma_n^{\boldsymbol{\theta}\mathbf{g}}-\mathbf{K}_n\frac{\mathrm{d} \Sigma_n^{\boldsymbol{\theta}\mathbf{g}}}{\mathrm{d}\mathbf{d}} \right],
   \end{aligned}
\end{equation}
where $\mathbf{K}_n=\Sigma^{\boldsymbol\theta \mathbf{g}}_n(\Sigma^{\mathbf{g}\mathbf{g}}_n+\Gamma)^{-1}$ is the Kalman Gain Matrix in the $n$-th EKI iteration. The gradient of $\mathbf{K}_n$, $\Sigma_n^{\boldsymbol{\theta}\mathbf{g}}$, and $\Sigma_n^{\mathbf{g}\mathbf{g}}$ with respect to $\mathbf{d}$ are written as
\begin{equation}
    \begin{aligned}\label{eq: EKI gradient 3}
    \frac{\mathrm{d} \mathbf{K}_n}{\mathrm{d} \mathbf{d}}&=\frac{\mathrm{d} \Sigma_n^{\boldsymbol{\theta}\mathbf{g}}}{\mathrm{d} \mathbf{d}} \left(\Sigma_n^{\mathbf{g}\mathbf{g}}+\Gamma\right)^{-1}- \Sigma_n^{\boldsymbol{\theta}\mathbf{g}} \left(\Sigma_n^{\mathbf{g}\mathbf{g}}+\Gamma\right)^{-1} \frac{\mathrm{d} \Sigma_n^{\mathbf{g}\mathbf{g}}}{\mathrm{d} \mathbf{d}}\left(\Sigma_n^{\mathbf{g}\mathbf{g}}+\Gamma\right)^{-1},\\
        \frac{\mathrm{d} \Sigma_n^{\boldsymbol{\theta}\mathbf{g}}}{\mathrm{d} \mathbf{d}}& = \frac{1}{J-1} \sum_{j=1}^J \left[\left(\frac{\mathrm{d} \bar{\boldsymbol\theta}_n}{\mathrm{d} \mathbf{d}} - \frac{\mathrm{d} \boldsymbol\theta_n^{(j)}}{\mathrm{d} \mathbf{d}}\right)^\top \left(\bar{\mathbf{g}}_n  - \mathbf{g}^{(j)}_n \right)+\left(\bar{\boldsymbol\theta}_n - \boldsymbol\theta_n^{(j)}\right)\left(\frac{\mathrm{d} \bar{\mathbf{g}}_n}{\mathrm{d} \mathbf{d}}  - \frac{\mathrm{d} \mathbf{g}^{(j)}_n}{\mathrm{d} \mathbf{d}} \right)^\top\right] ,\\
         \frac{\mathrm{d} \Sigma_n^{\mathbf{g}\mathbf{g}}}{\mathrm{d} \mathbf{d}}& = \frac{1}{J-1} \sum_{j=1}^J \left[\left(\frac{\mathrm{d} \bar{\mathbf{g}}_n}{\mathrm{d} \mathbf{d}}  - \frac{\mathrm{d} \mathbf{g}^{(j)}_n}{\mathrm{d} \mathbf{d}} \right)^\top\left(\bar{\mathbf{g}}_n  - \mathbf{g}^{(j)}_n \right)+\left(\bar{\mathbf{g}}_n  - \mathbf{g}^{(j)}_n \right)\left(\frac{\mathrm{d} \bar{\mathbf{g}}_n}{\mathrm{d} \mathbf{d}}  - \frac{\mathrm{d} \mathbf{g}^{(j)}_n}{\mathrm{d} \mathbf{d}} \right)^\top \right].
    \end{aligned}
\end{equation}

The mapping $\left[\mathbf{y},\mathbf{g}^{(j)}_{n-1}\right] \mapsto \tilde{D}_\text{KL}$ can be effectively handled by auto-differentiation tools such as PyTorch or JAX. The mapping $\mathbf{d} \mapsto \left[\mathbf{y},\mathbf{g}^{(j)}_{n-1}\right]$ involves the data measurement process and the forward model. In this work, we focus on the data measurement locations as the designs $\mathbf{d}$. If a spatially discretized solution of the full field is available, e.g., from the forward model output, interpolation can be performed to obtain the gradient information with respect to $\mathbf{d}$. In contrast, if the data is directly measured from the true field, infinitesimal perturbation analysis (IPA,~\cite{ho_perturbation_1983, huan_gradient-based_2014}) methods provide a feasible approximation. In practice, both the interpolation for the forward model output field and the perturbation-based sensitivity estimation in IPA can be directly approximated using the JAX built-in function \verb|jax.scipy.ndimage.map_coordinates|.

The primary computational complexity for the automatic differentiation of ensemble-based KL divergence arises from the iterative mechanism of EKI and the nested structure of the covariance matrix. First, the iterative update process in EKI leads to each ensemble in a given step being dependent on the previous step as shown in Eq.~\eqref{eq: eki updating}, thereby deepening the computational graph over successive iterations. Second, the covariance matrix exhibits a nested dependency structure, where a perturbation in a single element of the ensemble affects the entire matrix. Consequently, the computation graph grows increasingly intricate, as the gradient propagation must account for the interdependence of all ensemble members. These two factors significantly contribute to the computational burden of the automatic differentiation process. To quantify the impact of these complexities, it is essential to analyze the time complexity and peak memory cost associated with the automatic differentiation process.

The computational time complexity of auto-differentiating to the map $\mathbf{d} \mapsto \tilde{D}_\text{KL}$ remains of the same order as that of the forward mapping \cite{chen_autodifferentiable_2022} and scales linearly with the total number of EKI iterations $K$ at $\mathcal{O}(K)$, provided that the forward model of different ensemble members in EKI is simulated in parallel. Implementations of JAX loop functions like \verb|jax.scan| and \verb|jax.fori_loop| could help minimize the scaling factor as they directly compile an entire loop into a single XLA kernel without repeating Python interpretation. The computational time used for transferring data and commands between Python to XLA can be minimized from $2K$ times to twice.  An increase in time complexity is expected if the forward model of different ensemble members is executed sequentially or partially sequentially due to the memory limit.

The peak memory cost of auto-differentiating this mapping scales as $\mathcal{O}(KJ)$, further depends on the ensemble size $J$. By employing efficient memory management strategies, this memory requirement can be further reduced. For instance, checkpointing serves as an effective strategy to (i) further reduce memory consumption per iteration and (ii) lower the computational complexity from $\mathcal{O}(KJ)$ to $\mathcal{O}(J)$ by discarding intermediate results and recomputing them as needed during backpropagation, albeit at the cost of increased computational time.

While both AD-EnKF and our AD-EKI differentiate through Kalman updates, our work extends the auto-differentiable filtering paradigm from dynamic state estimation to static inverse problems and Bayesian experimental design. AD-EnKF is tailored to sequential data assimilation: it alternates forecast and analysis steps along a time series and differentiates through each assimilation cycle to improve real-time state. In contrast, our AD-EKI addresses static inverse problems and BED by repeatedly applying the EnKF analysis update conditioned on the same observations. For nonlinear forward models, a one-off Gaussian update yields only a local approximation of the parameter posterior~\cite{iglesias_ensemble_2013, schillings2017analysis}, which is inadequate for global design objectives. EKI overcomes this by iteratively refining the ensemble through multiple analysis updates, gradually capturing a reasonable posterior’s geometry~\cite{chada_iterative_2020, blomker_continuous_2021},  which facilitates information quantification and outer-loop design optimization.

\subsection{AD-EKI for A Differentiable Estimation of Expected Information Gain}\label{sec:AD_EKI_EIG}

$\tilde{D}_\text{KL}$ represents the actual information gain provided by some measurement data from the true system. In the context of experimental design, those measurement data are often unavailable a priori before an experiment is designed. Therefore, the expected information gain (EIG) is employed to design an experiment based on the modeled system and prior knowledge without requiring the actual data from the experiments. As expressed in Eq.~\ref{eq:EIG}, EIG integrates out the dependency on data $\mathbf{y}$ in the KL divergence $D_\text{KL}$ by the prior predictive distribution $p(\mathbf{y}|\mathbf{d})$. The differentiation formulae of $\tilde{D}_\text{KL}$ presented in Section~\ref{sec:AD_EKI_KL} remains valid, except for that data $\mathbf{y}$ is predicted by the model when EIG is employed as the utility function of BED.

Specifically, the distribution $p(\mathbf{y}|\mathbf{d})$ of the predicted data $\mathbf{y}$ is induced by the prior distribution $p(\boldsymbol{\theta})$ of the parameter $\boldsymbol{\theta}$, which is propagated through the PDE solution process under a given design $\mathbf{d}$ and subsequently transformed by the observation model with observing noise $\boldsymbol{\eta} \sim \mathcal{N}(\mathbf{0},\Gamma)$: $\{p(\boldsymbol{\theta}),\mathcal{N}(\mathbf{0},\Gamma)\} \mapsto p(\mathbf{y}|\mathbf{d})$. While directly evaluating this predictive distribution $p(\mathbf{y}|\mathbf{d})$ and calculating $\tilde{D}_\text{KL}(\mathbf{d,\mathbf{y}^{(m)}})$ for each sampled $\mathbf{y}^{(m)} \sim p(\mathbf{y}|\mathbf{d})$ is possible, this sampling process is not differentiable and reparameterization trick~\cite{kingma_auto-encoding_2013} is required. 

In practice, pairs of $\{\boldsymbol{\theta},\boldsymbol{\eta}\}^{m}$ can be directly sampled from prior $p(\boldsymbol{\theta})$ and Gaussian noise $\mathcal{N}(\mathbf{0},\Gamma)$. Therefore, samples of $\mathbf{y}^{m}$ can be obtained by $\mathbf{y}^{m}=G(\boldsymbol\theta^{m},\mathbf{d})+\boldsymbol{\eta}^{m}$, which is equivalent to sampling from $p(\mathbf{y}|\mathbf{d})$ but without accessing the distribution itself:
\begin{equation}
\label{eq:ensemble_EIG}
    \mathbb{E}_{\mathbf{y(\mathbf{d})}\sim p(\mathbf{y}|\mathbf{d})}[\tilde{D}_\text{KL}(\mathbf{d},\mathbf{y})] = \mathbb{E}_{\boldsymbol{\theta} \sim p(\boldsymbol{\theta}),\mathbf{\mathbf{\eta}}\sim p(\eta)} \{\tilde{D}_\text{KL}[\mathbf{d},\mathbf{y}(\boldsymbol{\theta}, \mathbf{\eta})]\}\approx\frac{1}{M}\sum_{m=1}^M \tilde{D}_\text{KL}[\mathbf{d},\mathbf{y}(\boldsymbol{\theta}^{m}, \mathbf{\eta}^{m})],
\end{equation}
where each $\tilde{D}_\text{KL}[\mathbf{d},\mathbf{y}(\boldsymbol{\theta}^{m}, \mathbf{\eta}^{m})]$ is evaluated by Eq.~\eqref{eq:ensemble_KLD} in this work. Consequently, differentiating EIG with respect to design $\mathbf{d}$ can be estimated as:
\begin{equation}
\begin{aligned}
    \frac{\mathrm{d} }{\mathrm{d} \mathbf{d}} \mathbb{E}_\mathbf{y(\mathbf{d})} [\tilde{D}_\text{KL}(\mathbf{d},\mathbf{y})] &\approx \frac{\mathrm{d} }{\mathrm{d} \mathbf{d}} \left[\frac{1}{M}\sum_{m=1}^M \tilde{D}_\text{KL}[\mathbf{d},\mathbf{y}(\boldsymbol{\theta}^{m}, \mathbf{\eta}^{m})]\right] \\
    &=  \frac{1}{M}\sum_{m=1}^M \frac{\mathrm{d} }{\mathrm{d} \mathbf{d}}\tilde{D}_\text{KL}[\mathbf{d},\mathbf{y}(\boldsymbol{\theta}^{m}, \mathbf{\eta}^{m})], 
\end{aligned}
\end{equation}
where the interchange of the order of differentiation and integration in the second equal is valid with mild assumptions that $\tilde{D}_\text{KL}$ and $\partial \tilde{D}_\text{KL}/\partial\mathbf{d}$ are continuous and an integrable function bounds the partial derivative term.

For a single sample of $\mathbf{y}^{m}$, the computational cost of AD-EKI remains nearly unchanged. As the calculation of $\tilde{D}_\text{KL}[\mathbf{d},\mathbf{y}(\boldsymbol{\theta}^{m}, \mathbf{\eta}^{m})]$ for each sample is independent, the total computational complexity for $M$ samples scales linearly as $\mathcal{O}(MJ)$. It is also worth noting that sequentially performing AD-EKI for each sample does not change the peak memory cost of $\mathcal{O}(J)$. The estimation of EIG in Eq.~\eqref{eq:ensemble_EIG} differs from a nested Monte Carlo method, which is commonly used in BED to approximate the double integral in EIG. The primary distinction between this formulation of Eq.~\eqref{eq:ensemble_EIG} and the nested Monte Carlo method lies in the estimation of $\tilde{D}_\text{KL}[\mathbf{d},\mathbf{y}(\boldsymbol{\theta}^{m}, \mathbf{\eta}^{m})]$, which is approximated by the ensemble Kalman inversion in this work. A key advantage is that a small ensemble size is often sufficient for ensemble Kalman methods, even if dealing with a high-dimensional problem, e.g., only an order of $10^2$ samples are typically required to handle parameter spaces with dimensions as high as $10^8$ in numerical weather prediction, with various techniques developed for a robust covariance estimation~\cite{tong_localization_2023,vishny_high-dimensional_2024}. 

\subsection{A Hybrid Framework for BED with Model Discrepancy}
\label{Sec: Hybrid Framework}

In the context of BED with model discrepancy, parameterizing the discrepancy in a data-driven way as Eq.~\eqref{eq:modeled_system} leads to two distinct sets of unknowns: the physical parameters $\boldsymbol{\theta}_\mathcal{G}$ and the network parameters $\boldsymbol{\theta}_\textrm{NN}$. Although performing BED on the joint parameter space $\{\boldsymbol{\theta}_\mathcal{G}, \boldsymbol{\theta}_\textrm{NN}\}$ is a natural first thought, incorporating network parameters significantly increases the dimensionality of the parameter space, posing a major challenge for full Bayesian methods due to computational intractability. It is worth noting that some recent works~\cite{wu2023large,wu2023fast,go2025sequential,go2025accurate} have been developed for Bayesian experimental design with high-dimensional unknown parameters, which rely on building an efficient and generalizable parameter-to-observable map and thus are not directly applicable for the modeled system in Eq.~\eqref{eq:modeled_system}. As an efficient alternative to full Bayesian methods, ensemble Kalman methods introduce Gaussian assumptions, are capable of handling high-dimensional parameter spaces, and demonstrate robust performance on nonlinear problems. However, the true distribution of physical parameters may be strongly non-Gaussian. Therefore, we propose a hybrid strategy that treats physical parameters with a full Bayesian approach, while dealing with neural network parameters via AD-EKI and gradient-based optimization. More specifically, we decompose the BED problem for $\{\boldsymbol{\theta}_\mathcal{G}, \boldsymbol{\theta}_\textrm{NN}\}$ into two sub-problems. First, we determine the optimal design for the physical parameters $\boldsymbol{\theta}_\mathcal{G}$ using standard BED techniques and update the belief distribution through Bayesian inference. Next, we determine the optimal experiment design for the calibration of neural network parameters $\boldsymbol{\theta}_\textrm{NN}$ by maximizing the ensemble-based utility function enabled by AD-EKI in Section~\ref{sec:AD_EKI_EIG}. With the data from the optimal design, we then update $\boldsymbol{\theta}_\textrm{NN}$ via a standard gradient-based optimization.

In this work, we focus on the scenario of sequential BED, where experiments are conducted sequentially for each stage, with the subsequent experiment in the subsequent stage benefiting from the information accumulated in previous ones. At each stage, we aim to identify two optimal designs, one for the low-dimensional unknown parameters for the physics-based model and the other for the high-dimensional unknown coefficients of the neural-network-based model discrepancy. 

At each stage of sequential BED, the optimal design for physical parameters is determined using standard BED: 
\begin{equation}
\label{eq:BED_optimization_G}
    \mathbf{d}^*_\mathcal{G} = \argmax_{\mathbf{d}\in\mathcal{D}} \mathbb{E}[D_{\textrm{KL}}(p(\boldsymbol{\theta}_\mathcal{G}|\mathbf{y};\boldsymbol{\theta}_\text{NN}) || p(\boldsymbol{\theta}_\mathcal{G})  )],
\end{equation}
where the posterior distribution of $\boldsymbol{\theta}_\mathcal{G}$ from the previous stage serves as the prior distribution $p(\boldsymbol{\theta}_\mathcal{G})$ for the current stage. Considering that $\boldsymbol{\theta}_\mathcal{G}$ is often low-dimensional, standard BED methods can be employed to solve the optimization problem in Eq.~\eqref{eq:BED_optimization_G}. With the optimal design $\mathbf{d}^*_\mathcal{G}$, the corresponding data $\mathbf{y}_\mathcal{G}$ is then used to update the belief of physical parameters by the Bayesian theorem conditioned on the current network coefficients $\boldsymbol\theta_\text{NN}$:
\begin{equation}
    p(\boldsymbol\theta_\mathcal{G}|\mathbf{y}_\mathcal{G},\mathbf{d}_\mathcal{G};\boldsymbol\theta_\text{NN})=\frac{p(\mathbf{y}_\mathcal{G}|\boldsymbol\theta_\mathcal{G},\mathbf{d}_\mathcal{G};\boldsymbol\theta_\text{NN})p(\boldsymbol\theta_\mathcal{G};\boldsymbol\theta_\text{NN})  }{p(\mathbf{y}_\mathcal{G}|\mathbf{d}_\mathcal{G};\boldsymbol\theta_\text{NN})},
\end{equation}
which also provides the MAP estimation of physical parameters $\boldsymbol\theta_\mathcal{G}^*$:
\begin{equation}
    \label{eq:selet 1 theta}\boldsymbol\theta_\mathcal{G}^*=\argmax_{\boldsymbol\theta_\mathcal{G}}
 \{ p(\boldsymbol\theta_\mathcal{G}|\mathbf{y}_\mathcal{G},\mathbf{d}_\mathcal{G};\boldsymbol\theta_\text{NN})\}.
\end{equation}

On the other hand, the optimal design for neural network coefficients is found by BED with the ensemble-based utility function:
\begin{equation}
    \mathbf{d}^*_\text{NN} = \argmax_{\mathbf{d}\in\mathcal{D}} \mathbb{E}[\tilde{D}_\text{KL}(p(\boldsymbol{\theta}_\text{NN}|\mathbf{y};\boldsymbol{\theta}_\mathcal{G}^*) || p(\boldsymbol{\theta}_\text{NN})  )],
\end{equation}
where $\tilde{D}_\text{KL}$ denotes the KL divergence estimated by the AD-EKI introduced in Section~\ref{sec:AD_EKI_KL}. For each stage of sequential BED with model correction, we assume a Gaussian prior over the neural network coefficients centered at their current values. With the optimal design $\mathbf{d}^*_\text{NN}$ being identified for the neural-network-based model discrepancy, the corresponding data $\mathbf{y}_\text{NN}$ is then used to update the network coefficients by a standard gradient-based optimization method. In practice, we leverage the entire accumulated dataset $\mathrm{Y}_\text{NN}^{(i-1)}=\{\mathbf{y}_\text{NN}^1,\mathbf{y}_\text{NN}^2,\cdots,\mathbf{y}_\text{NN}^{(i-1)}\}$ from previous stages in sequential BED, together with the data $\mathbf{y}_\text{NN}^{(i)}$ at the stage $i$ for the training of the neural-network-based model discrepancy.

A detailed algorithm for sequential BED with model discrepancy is summarized in Algorithm~\ref{alg:ActiveLearning}.

\begin{figure}[H]
    \centering
    \includegraphics[width=\linewidth]{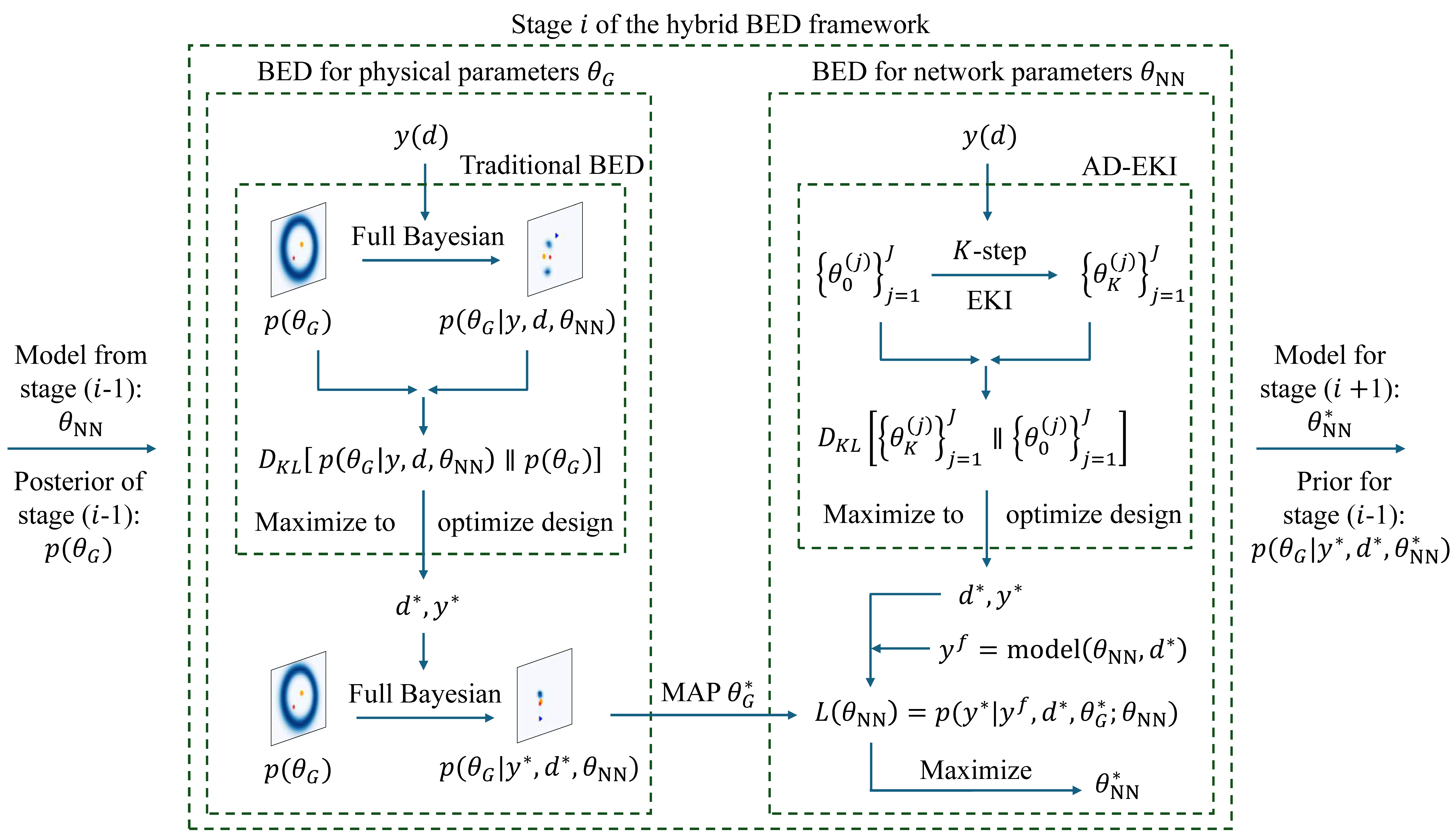}
    \caption{Schematic diagram of the hybrid BED framework: iterative learning of the parameters of a physics-based model and the model discrepancy.} 
    \label{fig:graphic_abstract}
\end{figure}

 \begin{algorithm}[H]
 \caption{Hybrid BED framework with Model Discrepancy via AD-EKI}
 \label{alg:ActiveLearning}
 \begin{algorithmic}
     \For {$i=1,2,..., N$} \Comment{Iterating sBED stage}
         \State \textbf{BED for physical parameter}
         \State $\mathbf{d}_\mathcal{G} \gets \arg \max_{\mathbf{d} \in \mathcal{D}}\mathbb{E}[U(\boldsymbol{\theta}_{\mathcal{G}},\mathbf{y},\mathbf{d}; \boldsymbol{\theta}_{\textrm{NN}})]$ \Comment{Traditional BED with $\boldsymbol{\theta}_{\textrm{NN}}$ fixed}
         \State $\mathbf{y}_\mathcal{G} \gets \mathbf{u}(\mathbf{d}_\mathcal{G}) + \boldsymbol{\epsilon}$ \Comment{Measurement at optimal design} 
         \State $\boldsymbol\theta_\mathcal{G} \gets \boldsymbol\theta_\mathcal{G}|\mathbf{y}_\mathcal{G}$ \Comment{Bayesian update}
         \State \textbf{BED for network parameter}
         \State $\boldsymbol\theta_\mathcal{G}^* \gets \argmax_{\boldsymbol\theta_\mathcal{G}} p(\boldsymbol{\theta}_\mathcal{G};\boldsymbol\theta_\text{NN})
         $ \Comment{MAP}
         \State $\mathbf{d}_{\textrm{NN}} \gets \arg \max_{\mathbf{d} \in \mathcal{D}}\mathbb{E}[\tilde{D}_{\textrm{KL}}(p(\boldsymbol{\theta}_{\textrm{NN}}|\mathbf{y}) || p(\boldsymbol{\theta}_{\textrm{NN}}) ;\boldsymbol{\theta}_\mathcal{G}^*)] $  \Comment{AD-EKI with $\boldsymbol{\theta}_\mathcal{G}^*$ fixed}
         \State $\mathbf{y}_{\textrm{NN}} \gets \mathbf{u}(\mathbf{d}_{\textrm{NN}}) + \boldsymbol{\epsilon}$ \Comment{Measurement at optimal design} 
         \State $\boldsymbol\theta_{\textrm{NN}} \gets \max_{\boldsymbol\theta_\text{NN}} L(\boldsymbol\theta_\text{NN};\boldsymbol\theta_\mathcal{G}^*,\mathbf{y}_\text{NN},\mathbf{d}_\text{NN})$ \Comment{Gradient-based optimization}
         \State $\boldsymbol\theta_\mathcal{G} \gets \boldsymbol\theta_\mathcal{G}|\mathbf{y}_\mathcal{G};\boldsymbol{\theta}_\text{NN}$ \Comment{Bayesian update conditioned on updated $\boldsymbol{\theta}_\text{NN}$}
     \EndFor
   
 \end{algorithmic}
 \end{algorithm}

\section{Numerical Results}
\label{sec: Numerical Results}

To demonstrate the performance of our proposed method, we study the contaminant source inversion problem, which is a classical example for BED and has been previously studied in \cite{huan_gradient-based_2014, shen_bayesian_2023, yang_active_2025}. The source inversion problem takes concentration measurements in a flow field governed by a convection-diffusion equation and then infers the plume source location. More specifically, the contaminant concentration $\mathbf{u}$ at a two-dimensional spatial location $\mathbf{z} = \{z_x, z_y\}$ and time $t$ is governed by the following equation: 

\begin{equation}
    \frac{\partial \mathbf{u}(\mathbf{z},t;\boldsymbol\theta)}{\partial t}=\nabla^2\mathbf{u}-v(t) \cdot \nabla \mathbf{u}+S(\mathbf{z},t;\boldsymbol\theta),~~~\mathbf{z} \in [z_L,z_R]^2,~~t>0
    \label{eq:true_system_example},
\end{equation}
where $v=\{v_x,v_y\} \in \mathbb{R}^2$ is a time-dependent convection velocity, $S$ denotes the source term with some parameters $\boldsymbol\theta$. In this work, the true system has an exponential decay source term in the following form with the parameters $\boldsymbol\theta=\{\theta_x,\theta_y,\theta_h,\theta_s\} \in \mathbb{R}^4$:
\begin{equation}
    S(\mathbf{z},t;\boldsymbol\theta)=\frac{\theta_s}{2\pi\theta_h^2}\exp \left(-\frac{(\theta_x-z_x)^2+(\theta_y-z_y)^2}{2\theta_h^2} \right),
    \label{eq:true_source_term}
\end{equation}
where \(\theta_x\) and \(\theta_y\) denote the source location, and \(\theta_h\) and \(\theta_s\) represent the source width and source strength. The initial condition is \(\mathbf{u}(\mathbf{z}, 0;\boldsymbol\theta) = 0\), and a homogeneous Neumann boundary condition is imposed for all sides of the square domain $[z_L,z_R]^2=[-2,3]^2$. 

\begin{figure}[H]
    \centering
    \includegraphics[width=0.93\linewidth]{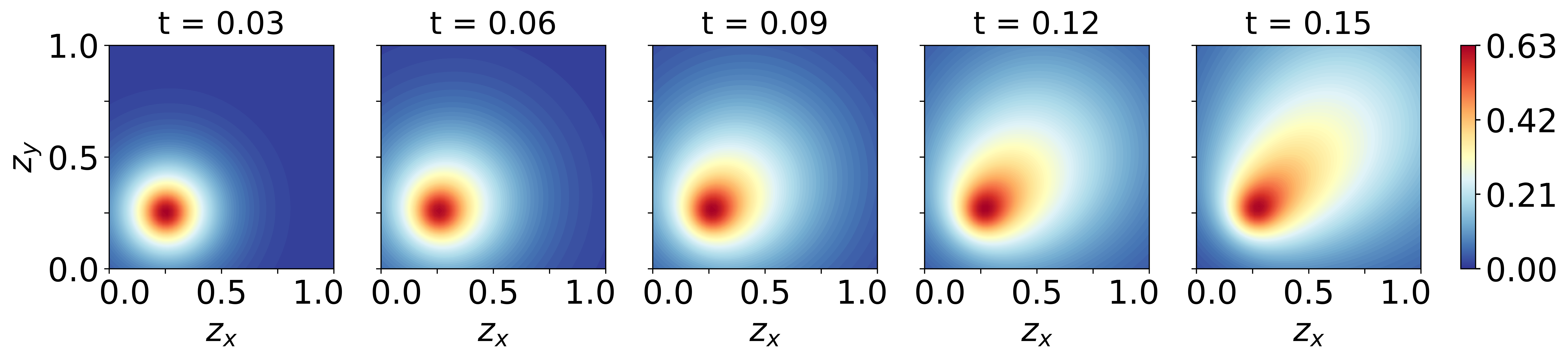}
    \caption{Concentration value at different times in a convection-diffusion field. The numerical simulation is performed in a larger domain ($[-2,3]^2$) and presented in a smaller one ($[0,1]^2$) to emphasize the areas close to the source location.}
    \label{PDE simulation results}
\end{figure}

\sloppy
In this work, the parameters in the source term are set as $\{\theta_x,\theta_y,\theta_h,\theta_s\}={0.25, 0.25, 2, 0.05}$ and convection velocity set as $v_x=v_y=50t$ and Fig.~\ref{PDE simulation results} presents the system state $\mathbf{u}$ calculated at $t=0.03,0.06,0.09,0.12,0.15$ time units, which illustrates how the source affects the concentration value at different locations across the domain through the convection and diffusion with the time evolution. The physical parameter support domain is set as $[0,1]^2$, which is smaller than the PDE solving domain to control computational cost and avoid boundary effect. Similar settings can also be found in~\cite{shen_bayesian_2023}.

The design $\mathbf{d}=\{d_x,d_y,d_t\}$ in this problem refers to the spatiotemporal coordinate to measure the concentration value. More specifically, each design involves the measurement of just one point in the domain. The spatial coordinates of the initial design start at $\{d_x^0,d_y^0\}=(0.5,0.5)$ and will gradually move around to other locations. The difference of spatial coordinates between two consecutive designs $(|d_x^{i+1} - d_x^{i}|, |d_y^{i+1} - d_y^{i}|)$ is constrained by the interval $[-0.2, 0.2]^2$ for physical parameters. The temporal coordinates of the measurement depend on different cases in general. At each time step, the posterior distribution from the previous stage serves as the prior for a new stage of BED as illustrated in Fig.~\ref{fig:graphic_abstract}. The current optimal design also serves as the starting point for the next stage of BED. In all numerical examples, the physics-based unknown parameters $\boldsymbol{\theta}_\mathcal{G}$ are the location $(\theta_x,\theta_y)$ of the source. The key motivations and findings of numerical results are summarized below:

\begin{itemize}
\item We first validate the performance of our proposed hybrid framework for BED in an example where the model error only exists as an incorrect value of the parameter $\theta_s$ set in the true source term of Eq.~\eqref{eq:true_source_term}. We demonstrate that the ensemble utility function based on AD-EKI identifies the most informative data and the iterative approach robustly achieves accurate estimation of $\boldsymbol{\theta}_\mathcal{G}$ and value of $\boldsymbol{\theta}_\text{NN}$. Detailed results can be found in Section~\ref{sec:Correct parametric error}.
\item We study a more challenging example where the model error exists as an incorrect knowledge of the function form for the source term, and such a model discrepancy is characterized by a neural network, which leads to high-dimensional unknown parameters $\boldsymbol{\theta}_\text{NN}$. In practice, the joint distribution of $\boldsymbol{\theta}_\mathcal{G}$ and $\boldsymbol{\theta}_\text{NN}$ is high-dimensional and the full BED would become expensive or even infeasible in such a challenging setup. We demonstrate that the proposed iterative approach provides an efficient and robust correction of structural model discrepancy and leads to a less biased estimation $\boldsymbol{\theta}_\mathcal{G}$. Detailed results can be found in Section~\ref{sec:Correct functional error}.
\item We investigate the scalability of the proposed hybrid framework for BED with respect to the ensemble size and the AD-EKI iterations. The numerical results confirm the discussions about the peak memory and the time cost in Sections~\ref{sec:AD_EKI_KL} and~\ref{sec:AD_EKI_EIG}. More detailed results can be found in Section~\ref{sec:results_scalability}.
\end{itemize}

\subsection{Parametric Model Error}\label{sec:Correct parametric error}

In this section, we demonstrate that the proposed approach in Section~\ref{Sec: Hybrid Framework} can efficiently and robustly calibrate the parametric model error, for which the standard BED problem is to identify optimal data for inferring the source location $\{\theta_x,\theta_y\}$. More specifically, we set up an example with the parametric model error where the true form of the source function in Eq.~\eqref{eq:true_source_term} is known but the value of $\theta_s$ is set incorrectly, e.g., due to the lack of knowledge on the strength of the source term. The true values are $\theta_x^\dagger=0.45$, $\theta_y^\dagger=0.25$, and $\theta_s^\dagger=2$. The advection velocity is assumed to be known and set as $v_x=v_y=20t$. The key motivation of this example is to validate the performance of the proposed method that identifies optimal designs for iteratively inferring the unknown parameters (i.e., source location for this example) of the physics-based model via standard BED and calibrating the model error (i.e., the value of the source magnitude). It is worth noting that the low-dimensional parametric model error allows directly targeting the joint distribution of all unknown parameters $\{\theta_x,\theta_y,\theta_s\}$ via standard BED with full Bayesian inference, and the results confirm that the proposed hybrid BED framework via AD-EKI can achieve similar performance.

The measurements are taken at $0.030,0.035,0.040,0.045,0.050,0.055$ time units as stages 1-6 of sequential BED. At the first stage, we employ a uniform prior distribution of $\{\theta_x, \theta_y\}$ over the range $[0,1]^2$, discretized into a $51\times 51$ grid of points, and set an initial value of $\theta_s = 3$. In each stage, standard BED is performed to infer the posterior distribution of $\{\theta_x, \theta_y\}$, and the optimal design for $\{\theta_x, \theta_y\}$ is determined accordingly based on the information gain introduced in Section~\ref{sec:OED}. At each stage (except for stage 1) after the distribution of $\{\theta_x, \theta_y\}$ is updated, optimal design for $\theta_s$ is determined based on the information gain estimated by AD-EKI, which was introduced in Section~\ref{sec:AD_EKI_KL}, to guide the data acquisition from the true system and facilitate the calibration of the model error $\theta_s$. The ensemble size is set as $30$, and the Gaussian distribution for the initial ensemble is assumed with the mean as the current value of $\theta_s$ and the variance as $1$.

\begin{figure}[H]
  \centering
  \includegraphics[width=0.95\linewidth]{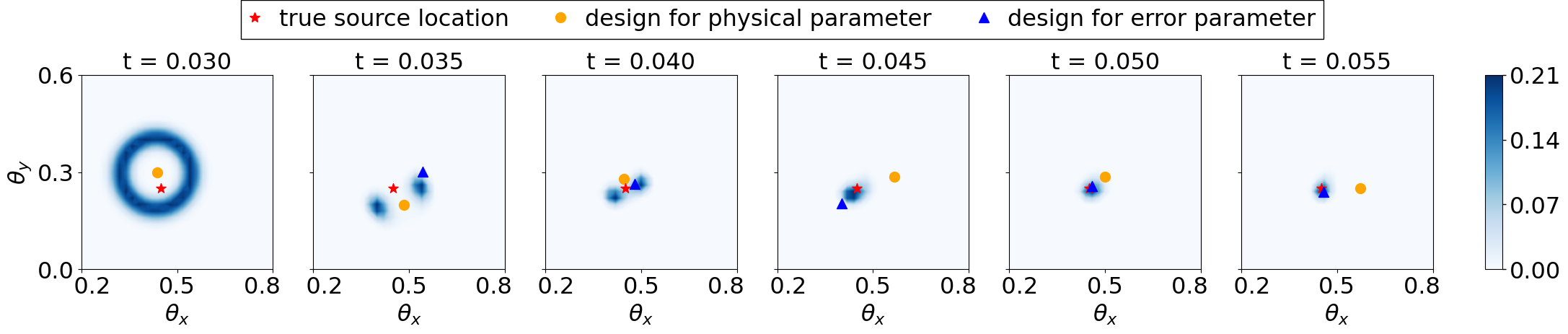}
  \caption{Posterior results of physical parameter $\{\theta_x,\theta_y\}$ via the hybrid approach on parametric error. The $\{\theta_x,\theta_y\}$ space is $[0,1]^2$ as before, with a zoomed-in view $[0.2,0.8]\times[0.0,0.6]$ to highlight detailed behaviors.}
  \label{fig: posterior of learning parameter error}
\end{figure}

Figure~\ref{fig: posterior of learning parameter error} shows the evolution of the posterior distribution for $\{\theta_x, \theta_y\}$ across six stages of sequential BED. It can be seen that the high probability regions gradually converge toward the true source location, which suggests an effective correction of parametric model error $\theta_s$. In addition, the high-probability regions not only shift toward the true source location but also become more concentrated, indicating an increasingly confident belief in the inferred source location. Notably, in the initial stages, the high-probability regions do not encompass the true source location, primarily due to the discrepancy introduced by the incorrect initial value of the model error parameter $\theta_s$. However, as the inference process advances, the posterior distribution incrementally adjusts, effectively mitigating the impact of the initial model discrepancy. This result demonstrates the capability of our framework to iteratively improve parameter estimation even in the presence of model discrepancy. 

\begin{figure}[H]
  \centering
  \begin{subfigure}[b]{0.4\textwidth}
    \centering
    \includegraphics[width=0.9\linewidth]{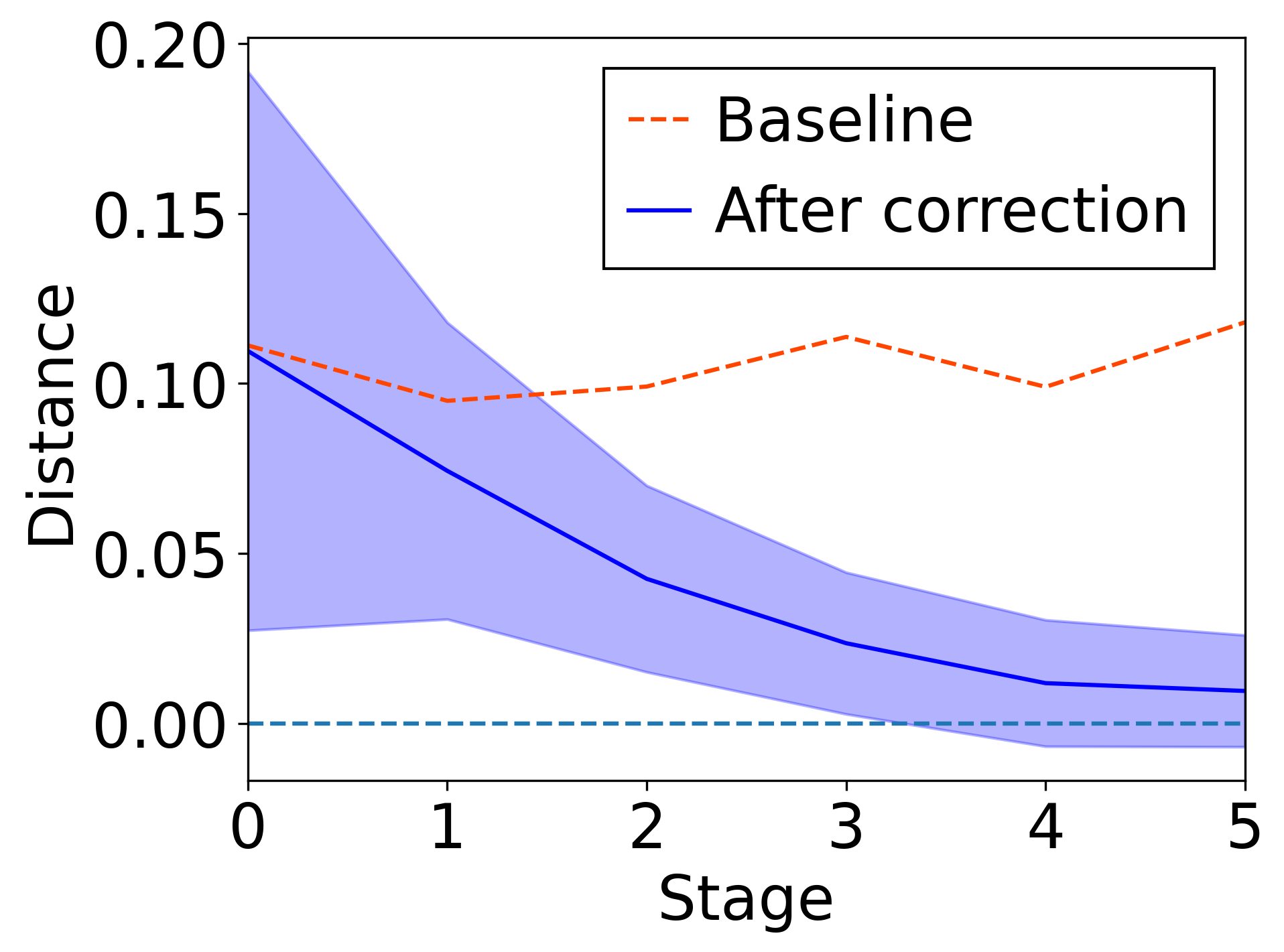}
    \caption{Distance between MAP and true center}
  \label{Estimation of physical parameter}
  \end{subfigure}%
  \hspace{3mm}
  \begin{subfigure}[b]{0.4\textwidth}
    \centering
    \includegraphics[width=0.9\linewidth]{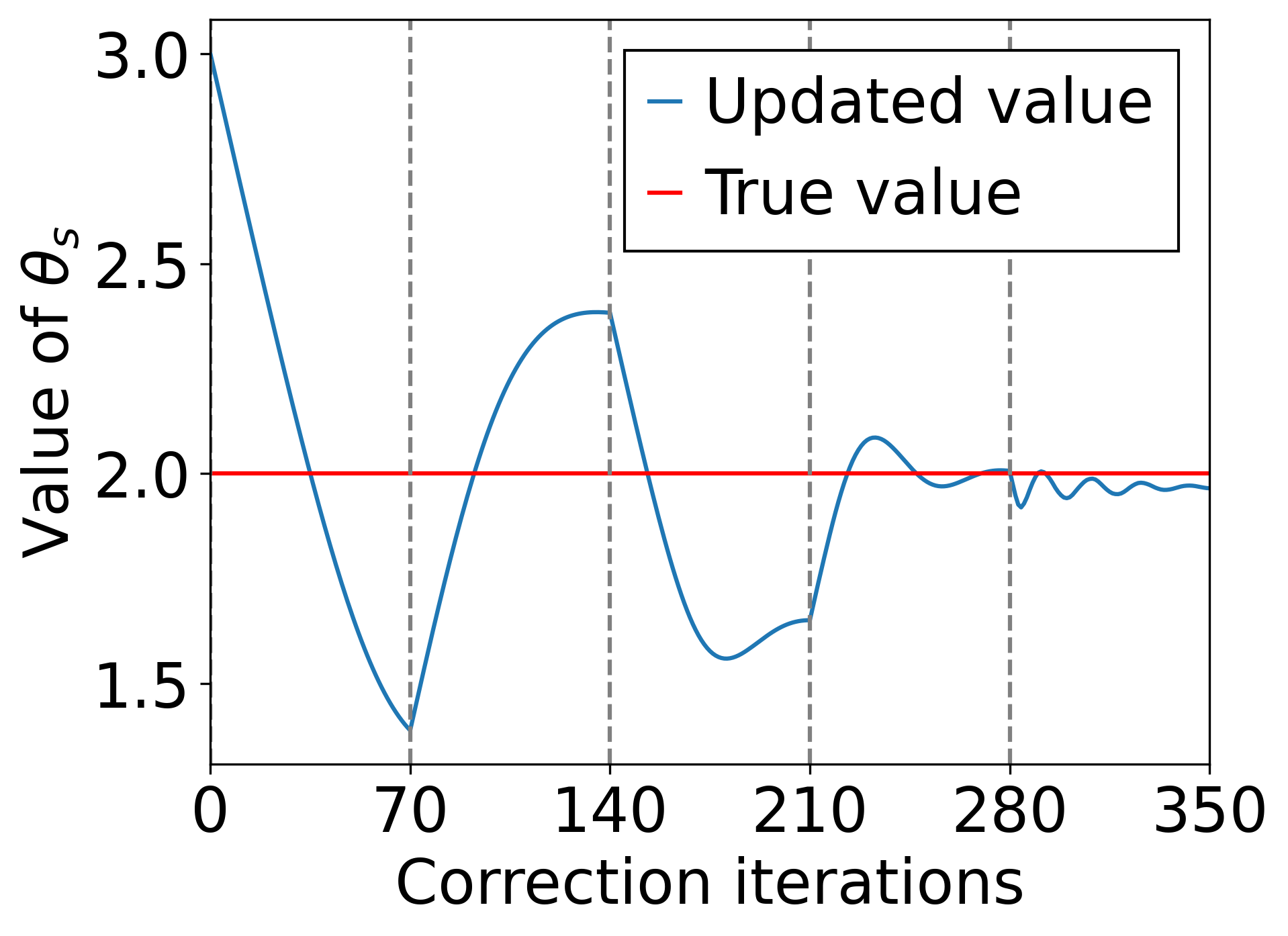}
    \caption{Updating results of error parameter}
  \label{Error parameter updating}
  \end{subfigure}%
  \caption{Results of the parametric error case: panel (a) shows a quantitative analysis of the posterior distribution of physical parameters, panel (b) shows the updating process of error parameter.}
  \label{parameter Results of learning parameter error}
\end{figure}

A quantitative metric to evaluate the performance of the posterior distribution of the source location $\{\theta_x, \theta_y\}$ is defined as the distance between the MAP point $\boldsymbol{\theta}_\mathcal{G}^*=\{\theta_x^*,\theta_y^*\}$ and the true value $\boldsymbol{\theta}_\mathcal{G}^\dagger=\{\theta_x^\dagger,\theta_y^\dagger\}$ in each stage:

\begin{equation}
\label{eq: distance}
    D = \|\boldsymbol{\theta}_\mathcal{G}^* - \boldsymbol{\theta}_\mathcal{G}^\dagger\|_2 =\sqrt{(\theta_x^*-\theta_x^\dagger)^2+(\theta_y^*-\theta_y^\dagger)^2},
\end{equation}
and the magnitude of uncertainty is quantified by:
\begin{equation}
\label{eq: uncertainty}
    \sigma_{\text{eq}} = \sqrt[4]{\lambda_1 \lambda_2} .
\end{equation}
where $\lambda_1$ and $\lambda_2$ are the eigenvalues of the posterior covariance matrix. Compared to Fig.~\ref{fig: posterior of learning parameter error}, Fig~\ref{Estimation of physical parameter} presents the performance of the inferred posterior distribution in a more quantitative way. The blue solid line represents the evolution of the distance metric $D$ across different stages, while the shaded region indicates the corresponding magnitude of uncertainty $\sigma_{\text{eq}}$. The overall decrease in distance confirms the improvement in parameter estimation accuracy.  

To further assess the impact of model discrepancy, we compare our proposed method with a baseline that does not account for error correction. The orange dashed line represents the estimation distance when the discrepancy parameter $\theta_s$ is not corrected. The significantly higher and less stable distance trajectory for this baseline case in Fig~\ref{Estimation of physical parameter} highlights the negative impact of ignoring the model discrepancy. This comparison demonstrates that our framework effectively mitigates the influence of model error, leading to improved parameter estimation performance for sequential BED with model discrepancy.

We also examine the results of corrected parametric model error $\theta_s$. As shown in Fig.~\ref{Error parameter updating}, the estimated value of $\theta_s$ gradually converges to the true value $\theta^\dagger_s = 2$ over correction iterations. This result confirms that the proposed framework effectively adjusts for model discrepancy, ensuring that the parametric error value aligns closely with the true underlying model. Compared to a previous work~\cite{yang_active_2025}, where the calibration of the model error parameter used optimal data from BED for physical parameters (i.e., source location in this example) and occasionally exhibited deviations from the true value, the estimation of model error parameter remains more robust here. This suggests that the optimal experimental design based on AD-EKI effectively identifies informative data directly for the model error parameter, ensuring a robust performance in the calibration of the error parameter $\theta_s$, which in turn benefits the inference of the physical parameters $\{\theta_x,\theta_y\}$.

\begin{figure}[H]
  \centering
  \begin{subfigure}[b]{0.45\textwidth}
    \centering
    \includegraphics[width=0.9\linewidth]{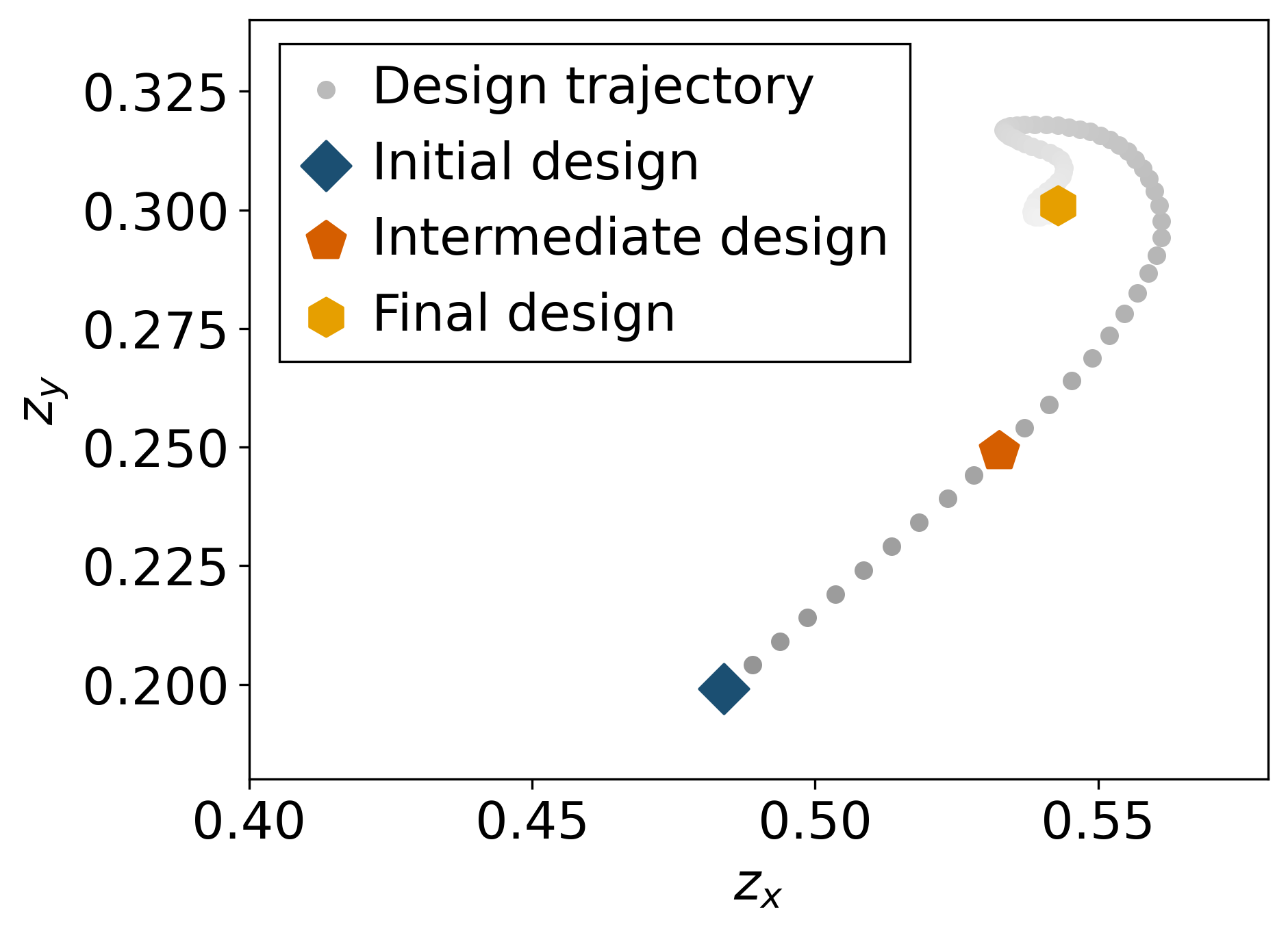}
    \caption{Design optimization trajectory}
  \label{design trajectory}
  \end{subfigure}
  \hspace{3mm}
  \begin{subfigure}[b]{0.45\textwidth}
    \centering
    \includegraphics[width=0.89\linewidth]{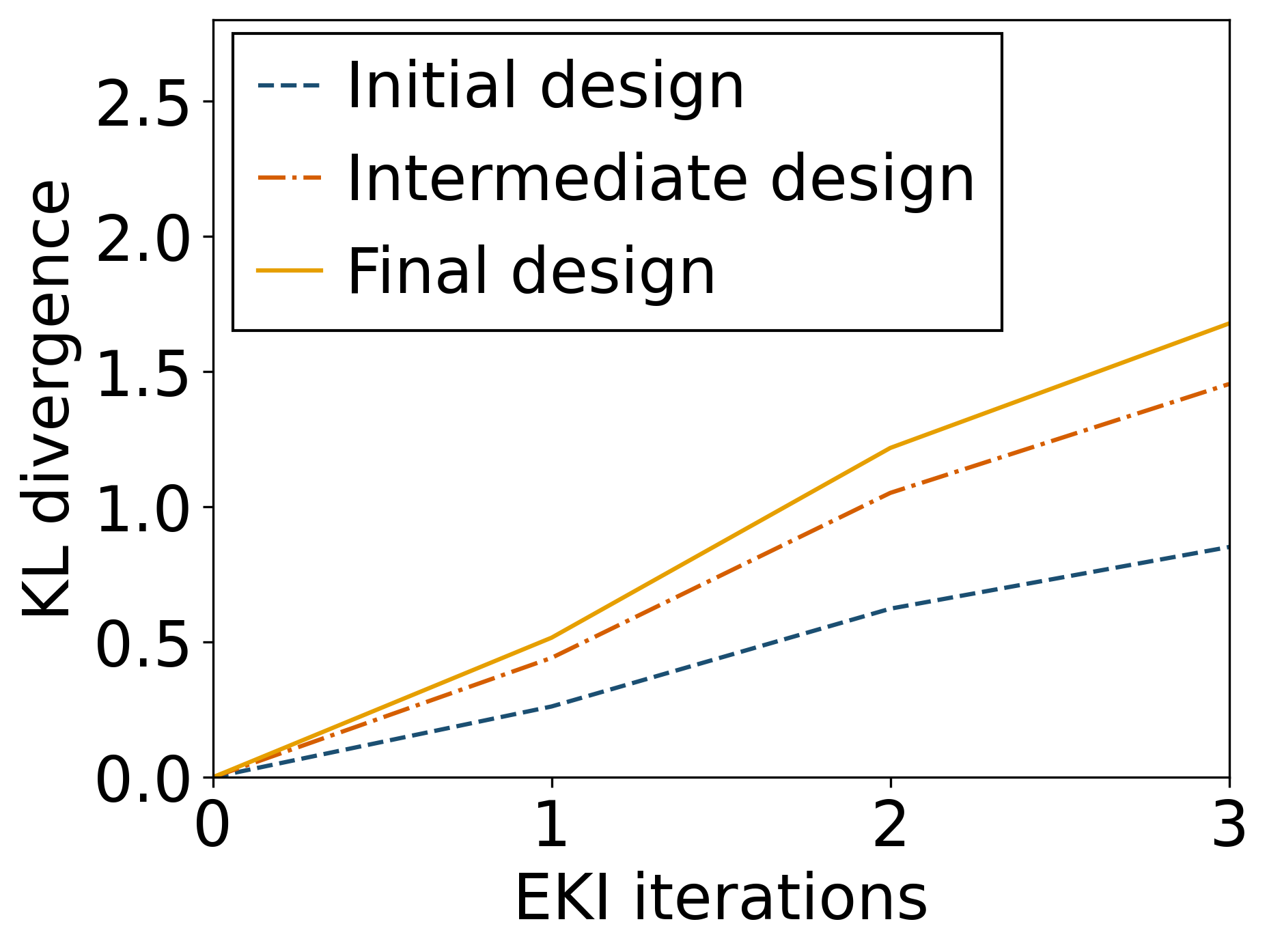}
    \caption{EKI details of different designs}
  \label{EKI updating}
  \end{subfigure}
  \caption{Design optimization results of the parametric error case: panel (a) shows the trajectory of design optimization via ensemble-based utility function and panel (b) shows the information gain at each EKI iteration of different designs. The gray dots in panel (a) represent the sequentially updated designs during the EKI iterations, with darker dots indicating earlier iterations and lighter dots indicating more recent iterations.}
  \label{parameter Results of learning parameter error}
\end{figure}

To further evaluate how the proposed framework enabled by AD-EKI distinguishes between designs of varying informativeness, we study the evolution of an AD-EKI approximated KL divergence during the first stage of sequential BED. A total of $70$ design updates were performed during this stage. Figure~\ref{design trajectory} presents three designs: (i) the initial design, (ii) an intermediate ($10$-th) design, and (iii) the final design, selected along the search trajectory for the optimal design. The AD-EKI approximated KL divergence in Fig.~\ref{EKI updating} is based on the average results obtained from 50 independent runs with randomly initialized ensembles. For each design, the increasing KL divergence along the iteration steps reflects the nature of EKI as an iterative updating process, where information from the data is progressively incorporated into the updated ensemble. Moreover, the differences between the three curves at each iteration step confirm that the final design leads to the largest approximated KL divergence and thus is considered as most informative. 

\subsection{Structural Model Error}
\label{sec:Correct functional error}

In this section, we focus on the scenario that structural error exists in the modeled system of Eq.~\eqref{eq:modeled_system}, i.e., $\mathcal{G}^\dagger - \mathcal{G}$ is non-negligible. The structural model error is characterized by a neural-network-based model as shown in Eq.~\eqref{eq:modeled_system} with parameters $\boldsymbol{\theta}_\text{NN}$ in a high-dimensional space. This scenario is challenging for the BED methods with full Bayesian inference, which often becomes prohibitively expensive or even infeasible due to the computational cost of handling high-dimensional parameter spaces. Our goal is to demonstrate that the proposed method provides an efficient and robust calibration of high-dimensional unknown parameters for structural model discrepancy based on the optimal designs identified by the hybrid BED framework via AD-EKI.

The modeled system is still in the same form of Eq.~\eqref{eq:true_system_example}, while the detailed form of the source term is defined by:
\begin{equation}
    S(\mathbf{z},t;\bm\theta)=\frac{3\theta_s}{\pi\left(\frac{(\theta_x-z_x)^2+(\theta_y-z_y)^2}{2\theta_h^2}+2\theta_h^2\right)},
    \label{eq: c sourse}
\end{equation}
which differs from the exponentially decaying source term as defined in Eq.~\eqref{eq:true_source_term} for the true system. Without addressing this model structural error, the inference of source location through standard BED yields biased results. In this work, a neural network $\mathbf{NN}(z_x,z_y,\theta_x,\theta_y;\boldsymbol{\theta_\text{NN}})$ is employed to characterize the model structural error. We utilize a fully connected neural network consisting of 37 parameters, a configuration that introduces a higher-dimensional parameter space than the example of parametric model error in Section~\ref{sec:Correct parametric error}. The neural network takes a 2-dimensional input vector, representing the relative distance from a specific grid point to the source location. The output is a scalar value for this point, indicating the discrepancy between the true and modeled source term values. The fully connected network is designed with two hidden layers, each containing four neurons, and employs a tanh activation function for both layers. The measurements are taken at $0.030,0.035,0.040,0.045,0.050,0.055$ time units as stages 1-6 of sequential BED. We employ a uniform prior distribution over the range $[0,1]^2$ for $\{\theta_x, \theta_y\}$, discretized into a $51\times 51$ grid, with the true value of the source location set at $\{0.25,0.25\}$. The values of $\{\theta_s^\dagger, \theta_h^\dagger\}$ are set as $\{2, 0.05\}$ in both the true system and the modeled one. The advection velocity is known and set as $v_x=v_y=50t$. The ensemble size for network parameters is set to $40$. Each element of $\boldsymbol{\theta}_\text{NN}$ is assumed to have a zero-mean Gaussian error with a variance of $0.09$ for the prior distribution.

\begin{figure}[H]
  \centering
  \begin{subfigure}[b]{1\textwidth}
    \centering
    \includegraphics[width=0.95\linewidth]{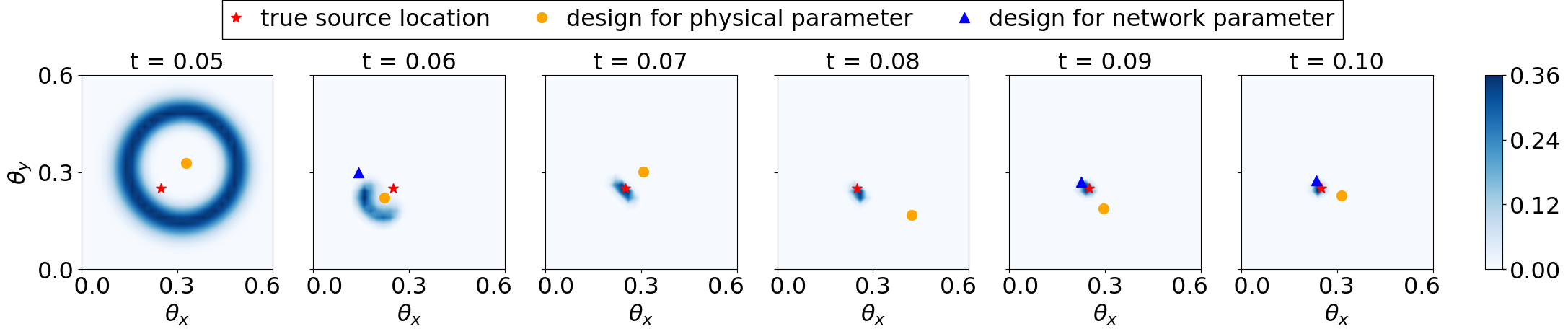}
    \caption{Using measured data}
    \label{fig:posterior_DP_1}
  \end{subfigure}%
  \vspace{10pt}
  \begin{subfigure}[b]{1\textwidth}
    \centering
    \includegraphics[width=0.95\linewidth]{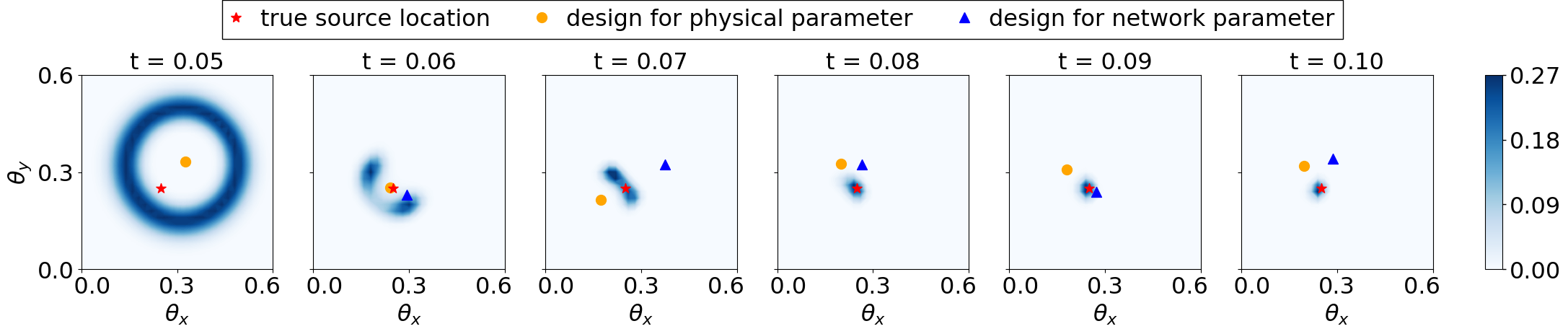}
    \caption{Using predicted data}
    \label{fig:posterior_DP_2}
  \end{subfigure}%
  \caption{Posterior distributions of inferred source location $\{\theta_x,\theta_y\}$ in BED: based on (a) measured data and (b) predicted data.}
  \label{fig: correct network error}
\end{figure}

\begin{figure}[H]
    \centering
    \includegraphics[width=0.4\linewidth]{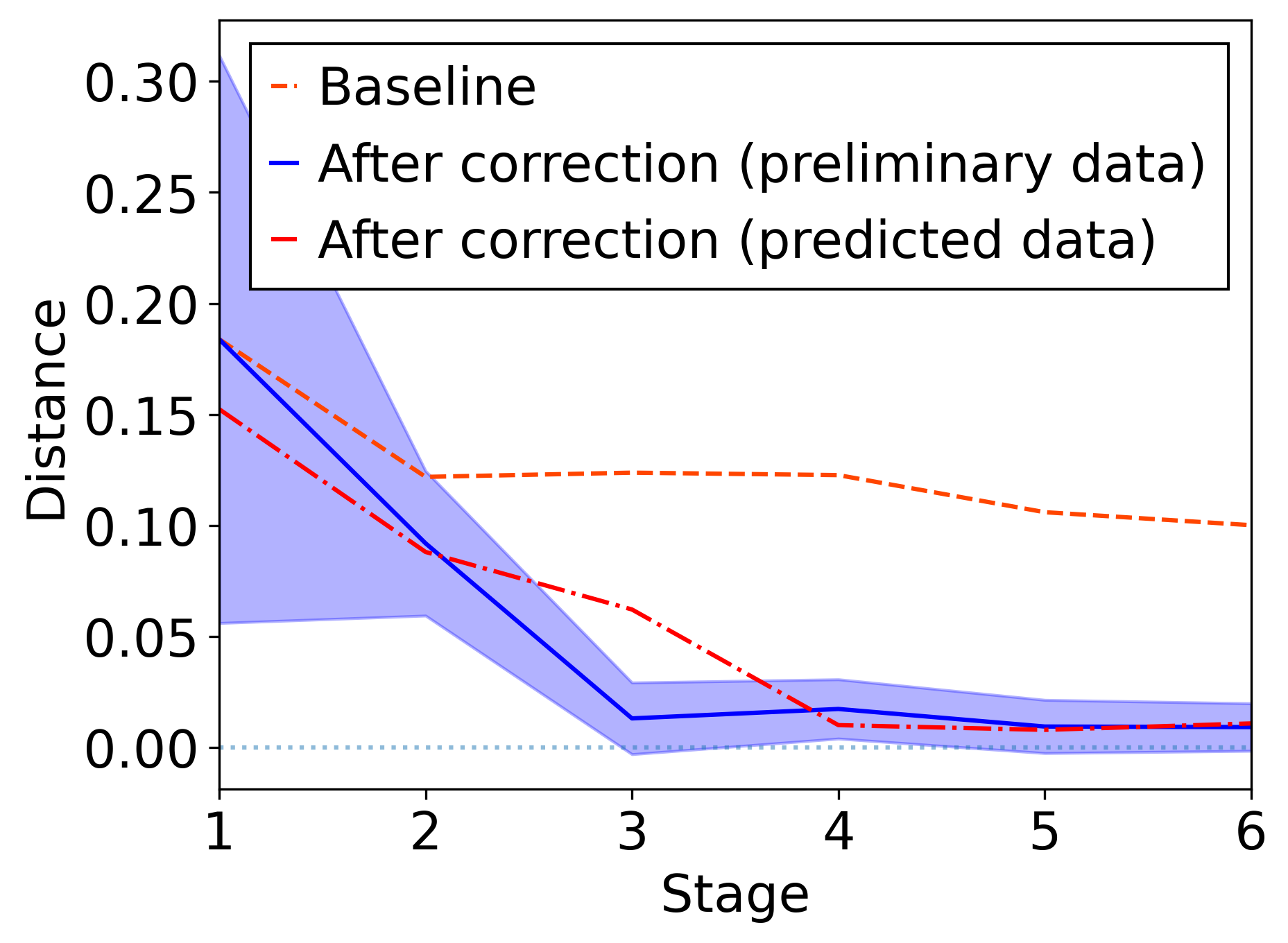}
    \caption{Posterior distribution of physical parameters: quantitative analysis.}
    \label{fig:nn_case_distance_with_predicted}
\end{figure}

We first present the posterior results of physical parameters (i.e., the source location) using measured data and predicted data in the evaluation of the utility function for finding an optimal design. The high probability area in both panels of Fig.~\ref{fig: correct network error} gradually converges to the true source location. Nevertheless, the results of using predicted data show a slower convergence rate than those using measured data, which is mainly because introducing information from the true system would mitigate the overestimation of the design selected by EIG relative to its true information gain. In many real-world applications, measurement data from the true system may be unavailable in the process of finding an optimal design, and it is worth noting that the results with predicted data become comparable to the ones using the true measurement data after a few stages, which confirms the satisfactory performance of the proposed method in a more realistic setting. The distance metric in Fig.~\ref{fig:nn_case_distance_with_predicted} quantitatively illustrates the convergence trend of the inferred posterior distribution of the source location to the true values. The baseline in Fig.~\ref{fig:nn_case_distance_with_predicted} corresponds to the results of standard BED that ignores the model discrepancy. The improved results of the proposed method, whether using the true measurement data or the predicted data for finding an optimal design, confirm that the hybrid BED framework via AD-EKI can well handle model discrepancy and provide a more robust estimation of parameters in the physics-based model.

\begin{figure}[H]
  \centering
  \begin{subfigure}[b]{0.45\textwidth}
    \centering
    \includegraphics[width=0.9\linewidth]{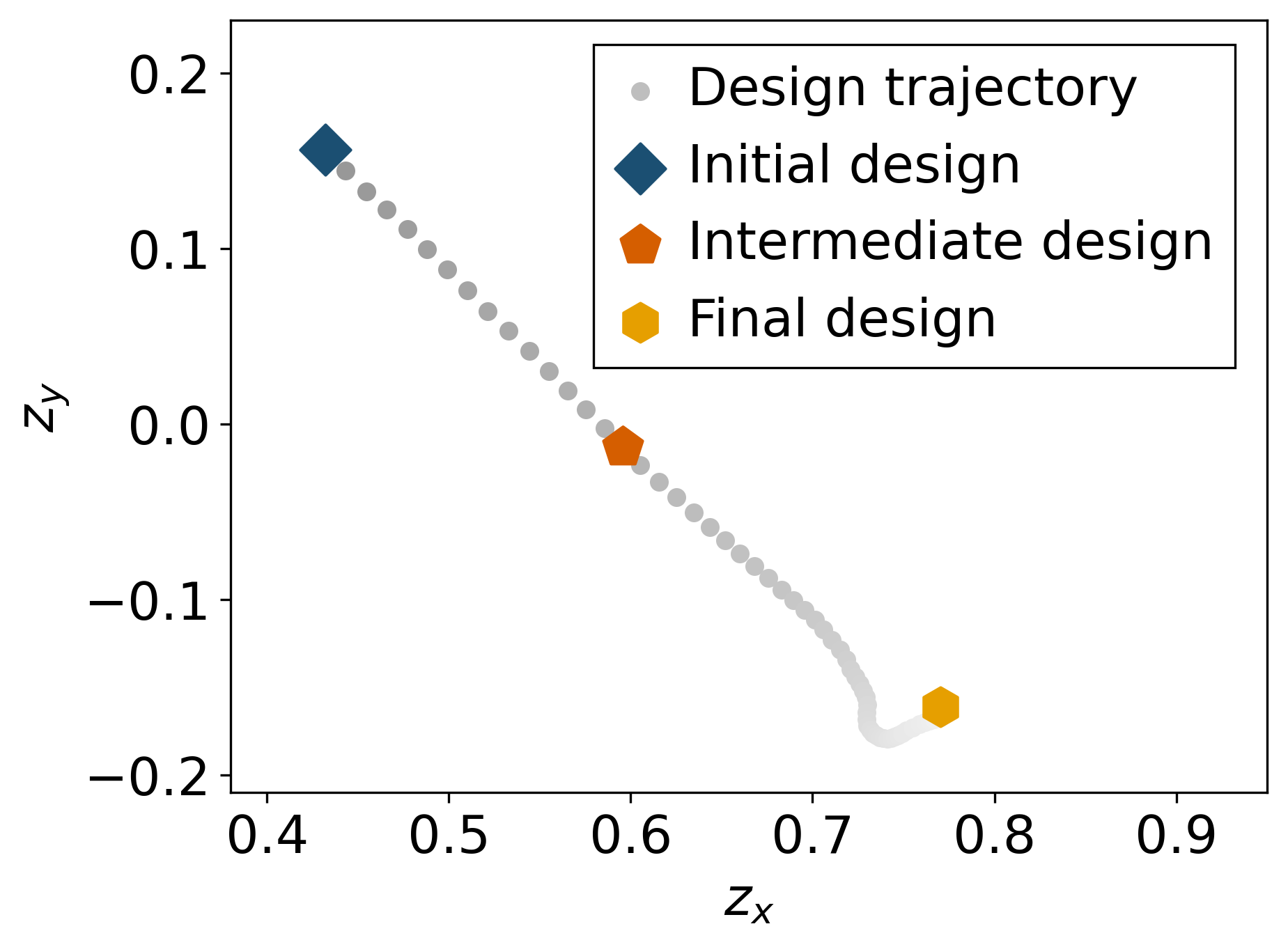}
    \caption{Design optimization trajectory}
  \label{fig: Design optimization trajectory}
  \end{subfigure}
  \hspace{3mm}
  \begin{subfigure}[b]{0.45\textwidth}
    \centering
    \includegraphics[width=0.89\linewidth]{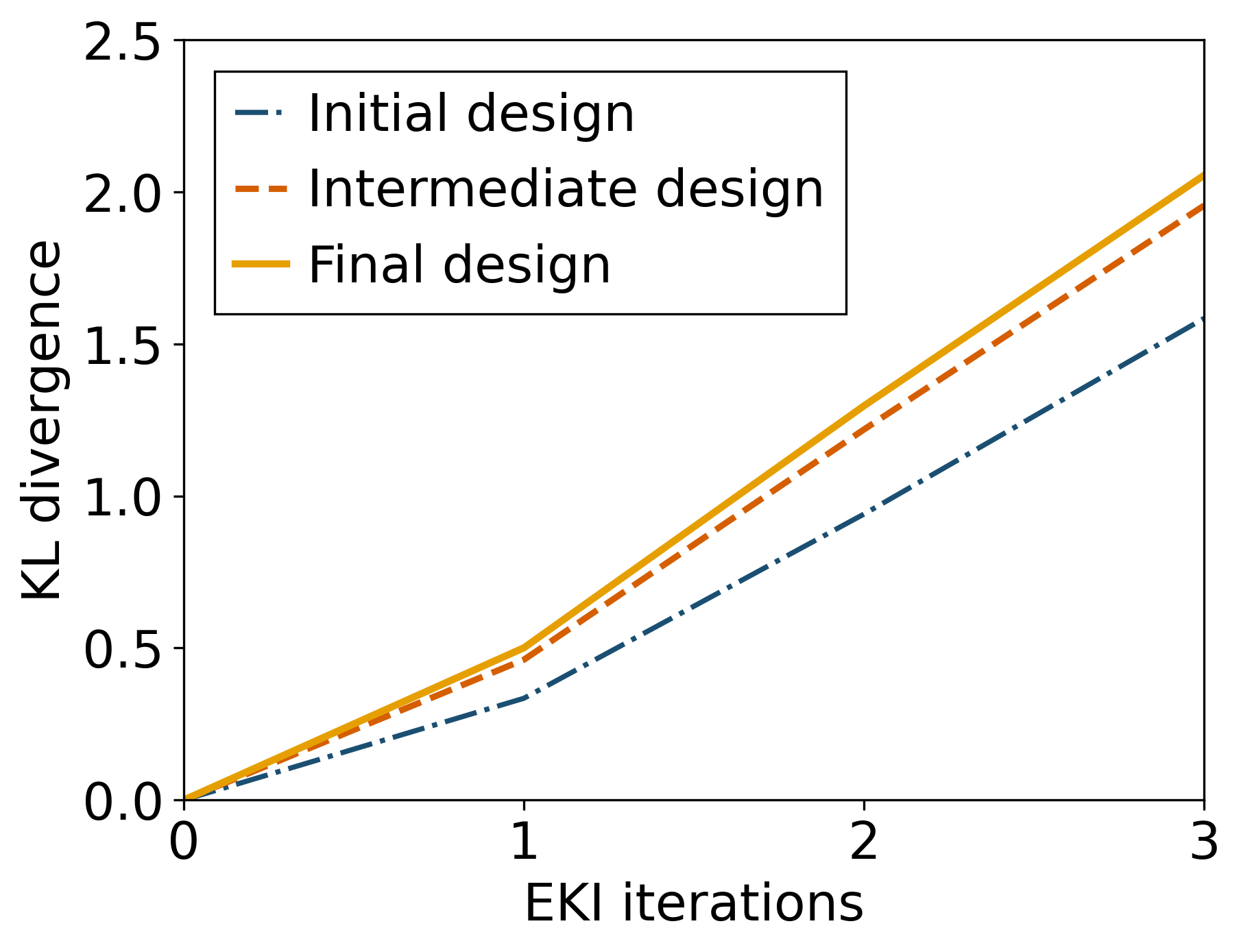}
    \caption{EKI details of different designs}
  \label{fig: EKI details of different designs}
  \end{subfigure}
  \caption{Design optimization results of the structural error case: panel (a) shows the trajectory of design optimization via ensemble-based utility function and panel (b) shows the information gain at each EKI iteration of different designs.}
  \label{parameter_results_of_learning_parameter_error}
\end{figure}

Figure~\ref{fig: EKI details of different designs} presents the information gain at each iteration of the AD-EKI for three design stages: (i) the initial design, (ii) an intermediate design, and (iii) the final design as shown in Fig.~\ref{fig: Design optimization trajectory}. In total, $64$ design updates were performed during the EKI process. Although the initial design optimization was set for $70$ steps, the process was automatically terminated at the $64$-$th$ step due to reaching the predefined design search boundaries, triggering the stopping criterion. The $15$-$th$ design has a physical location approximately halfway between the initial and final designs and is selected to illustrate intermediate results of information gain over the optimization. The information gain of the updated designs consistently exceeds that of the preceding ones at each EKI iteration, which confirms the effectiveness of the ensemble-based information gain as a criterion for guiding the design optimization procedure.

We further study some quantitative metrics on the accuracy of the solution field from the modeled system and summarize the results in Table~\ref{tab: nn case 1}, to demonstrate the effect of correcting the model discrepancy. More specifically, we calculate the mean square error and relative error as below:
\begin{equation}
    \begin{aligned}
        \text{MSE}&=\frac{1}{N}\sum_{z_x,z_y} (\mathbf{u}(z_x,z_y) - \mathbf{u}^\dagger(z_x,z_y))^2,\\
        \text{RE}&=\frac{\sum_{z_x,z_y} |\mathbf{u}(z_x,z_y) - \mathbf{u}^\dagger(z_x,z_y)|}{\sum_{z_x,z_y}  |\mathbf{u}^\dagger(z_x,z_y)| },
    \end{aligned}
\end{equation}
where $\mathbf{u}$ is the solution field from the modeled system and $\mathbf{u}^\dagger$ is the one from the true system. We first present the total error across the entire domain (i.e., $[0,1]^2$), which provides a general evaluation of the correction performance. Additionally, we show the local error in a smaller region (i.e., $\pm 0.04$ in both x and y directions) near the design for the physical parameters in the current stage, as an accurate estimation in this region is crucial for improving the Bayesian updating of the physical parameters. Since the model passed to the next stage is immediately used for design optimization, we also evaluate the effect of model correction on the subsequent stage. Specifically, we use the corrected model to optimize the next-stage design (for the Bayesian inference of physical parameters, i.e., the source location in this example) and compute the local error near the optimal design, comparing it to the results obtained using the uncorrected model.

\begin{table}[H]
    \centering
    \caption{Mean squared error (MSE) and relative error (RE) of the solution field from the modeled system.}
    \label{tab: nn case 1}
    \resizebox{\textwidth}{!}{
    \begin{tabular}{|c|c|c|c|c|c|c|c|c|c|c|}
    \hline
    Stage Index & \multicolumn{2}{|c|}{2} & \multicolumn{2}{|c|}{3} & \multicolumn{2}{|c|}{4} & \multicolumn{2}{|c|}{5} & \multicolumn{2}{|c|}{6}\\ \hline
    Model Correction & No & Yes & No & Yes & No & Yes & No & Yes & No & Yes \\ \hline
    \multicolumn{11}{|c|}{Effect on the current stage}\\ \hline
    Total MSE       & 0.3049  & 0.5949  & 0.8114 & 0.2447  & 0.3468 & 0.1012 & 0.1066 & 0.1311 & 0.1590 & 0.2240    \\ \hline
    Total RE        & 0.5348  & 0.8899  & 0.9437 & 0.5190  & 0.5502 & 0.3021   & 0.2837 & 0.3668 & 0.3733 & 0.4542 \\ \hline
    Local MSE       & 0.1915  & 0.1047  & 0.0503 & 0.0016  & 0.0560 & 0.0200   & 0.0477 & 0.0014 & 0.0024 & 0.0007 \\ \hline
    Local RE        & 0.4815  & 0.3489  & 0.2245 & 0.0340  & 0.4674 & 0.2774   & 0.2440 & 0.0372 & 0.0396 & 0.0220 \\ \hline
    \multicolumn{11}{|c|}{Effect on the next stage}\\ \hline
    Local MSE       & 0.0462 & 0.0503  & 0.1952 & 0.0569   & 0.1032 & 0.0477 & 0.0465 & 0.0024 & -  & -  \\ \hline
    Local RE        & 0.5556 & 0.2245  & 0.5089 & 0.4674   & 0.5214 & 0.2440 & 0.2361 & 0.0396 & -  & -  \\ \hline
    \end{tabular}
    }
\end{table}

Table~\ref{tab: nn case 1} presents the error analysis of the solution field from the modeled system with and without the model discrepancy correction. It can be seen that the errors in a local region near the current optimal design are consistently smaller for the results with model correction, indicating the key merit of the proposed hybrid BED framework that identifies optimal designs to better characterize the model discrepancy. Compared to the local errors, the errors across the whole domain may not always decrease with the model correction, which is expected since the correction is based on a limited number of data points, and regions lacking data remain unconstrained due to the absence of information. Considering that the Bayesian inference of physical parameters is mostly sensitive with respect to the model discrepancy in the regions near the optimal designs at each stage, the proposed framework is sufficient to achieve satisfactory results of inferred physical parameters and demands much less data (i.e., expensive experiments in many real-world applications), compared to empirically choosing data for the calibration of model discrepancy.

\subsection{Empirical Results on Scalability of AD-EKI}
\label{sec:results_scalability}
\begin{figure}[H]
  \centering
  \begin{subfigure}[b]{0.45\textwidth}
    \centering
    \includegraphics[width=0.9\linewidth]{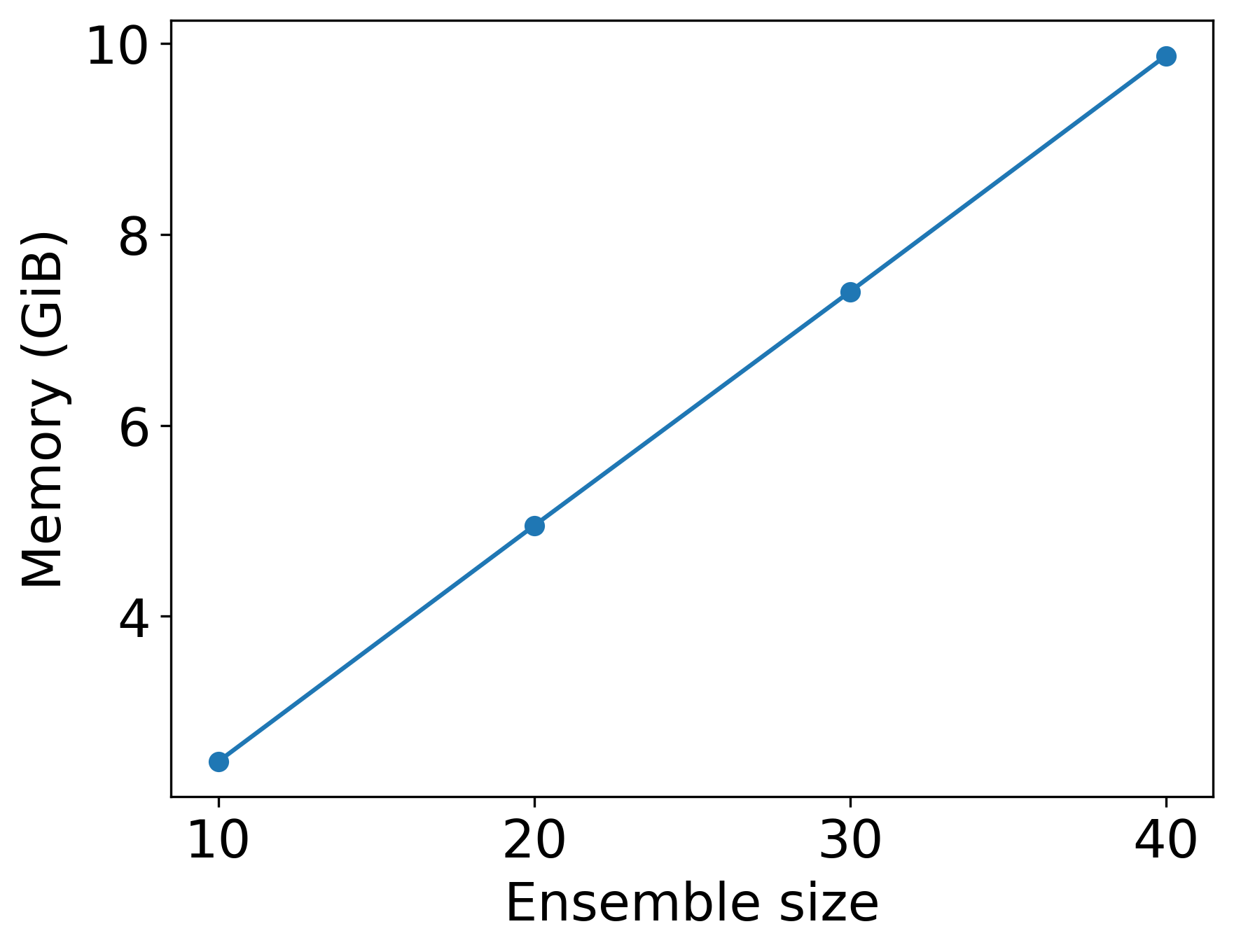}
    \caption{Memory cost}
  \label{fig: memory cost size}
  \end{subfigure}
  \hspace{3mm}
  \begin{subfigure}[b]{0.45\textwidth}
    \centering
    \includegraphics[width=0.9\linewidth]{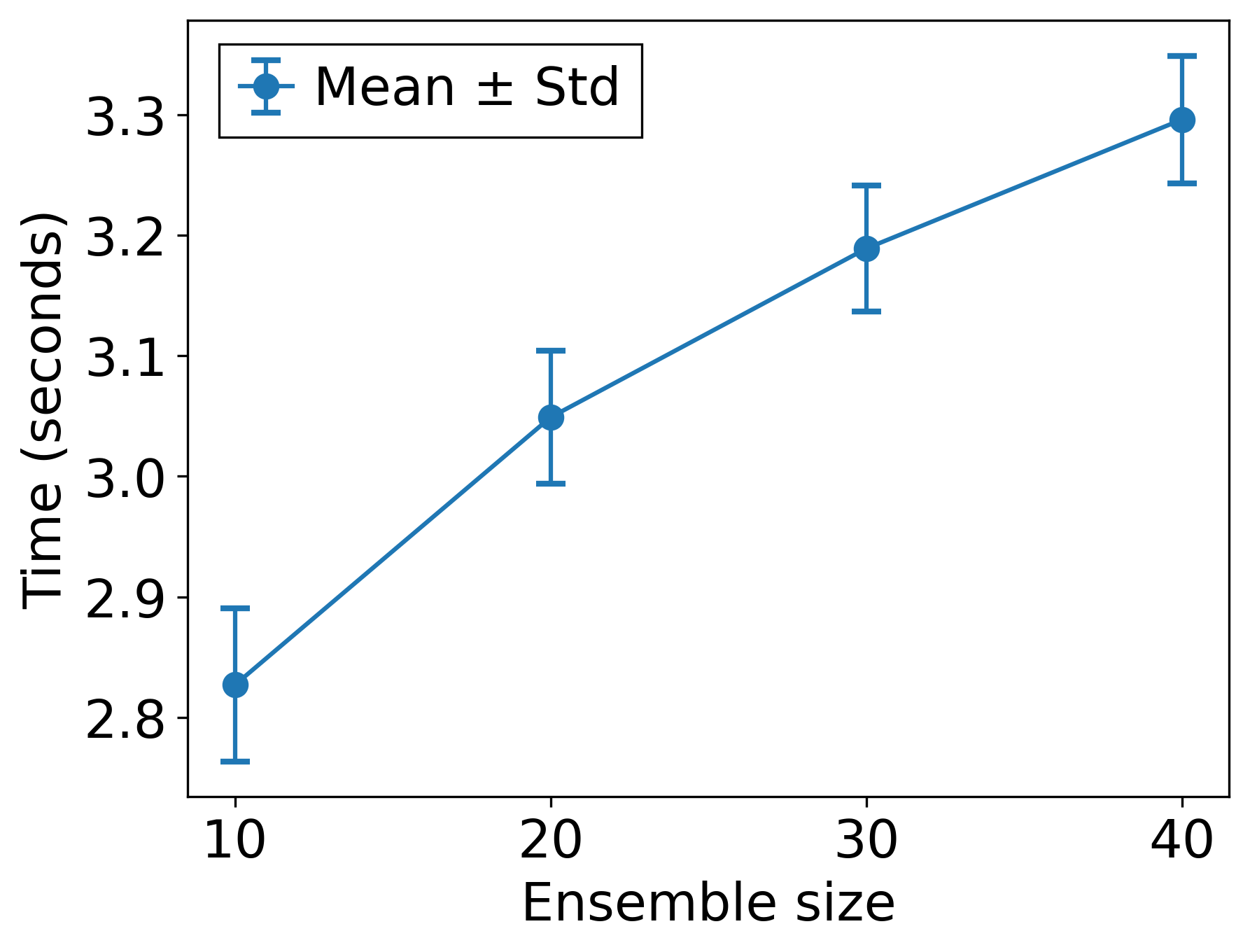}
    \caption{Time cost}
  \label{fig: Time cost size}
  \end{subfigure}
  \caption{Comparison of computation cost under different strategies across different numbers of ensemble size.}
  \label{fig: cost - ensemble size}
\end{figure}

\begin{figure}[H]
  \centering
  \begin{subfigure}[b]{0.45\textwidth}
    \centering
    \includegraphics[width=0.9\linewidth]{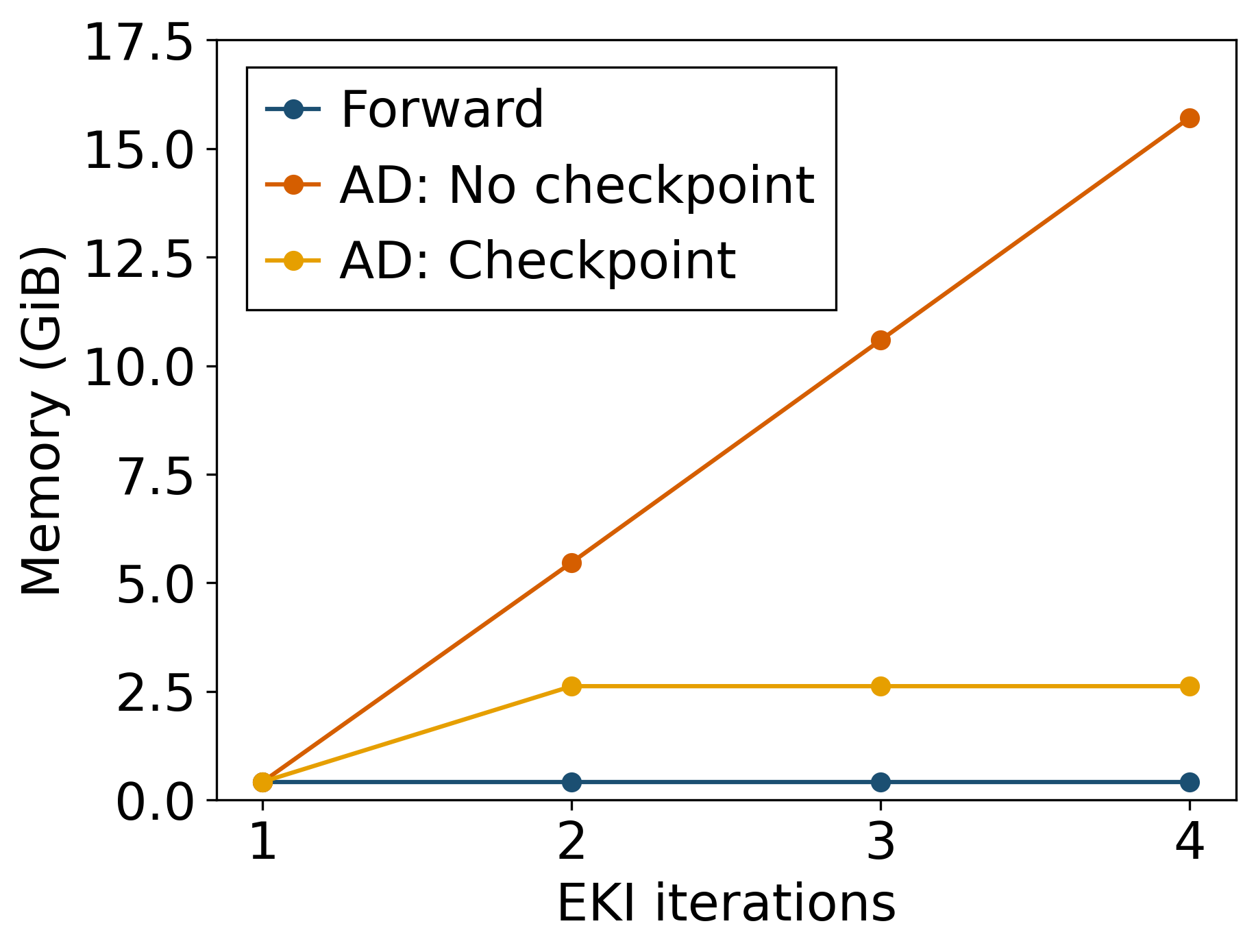}
    \caption{Memory cost}
  \label{fig: memory cost step}
  \end{subfigure}
  \hspace{3mm}
  \begin{subfigure}[b]{0.45\textwidth}
    \centering
    \includegraphics[width=0.85\linewidth]{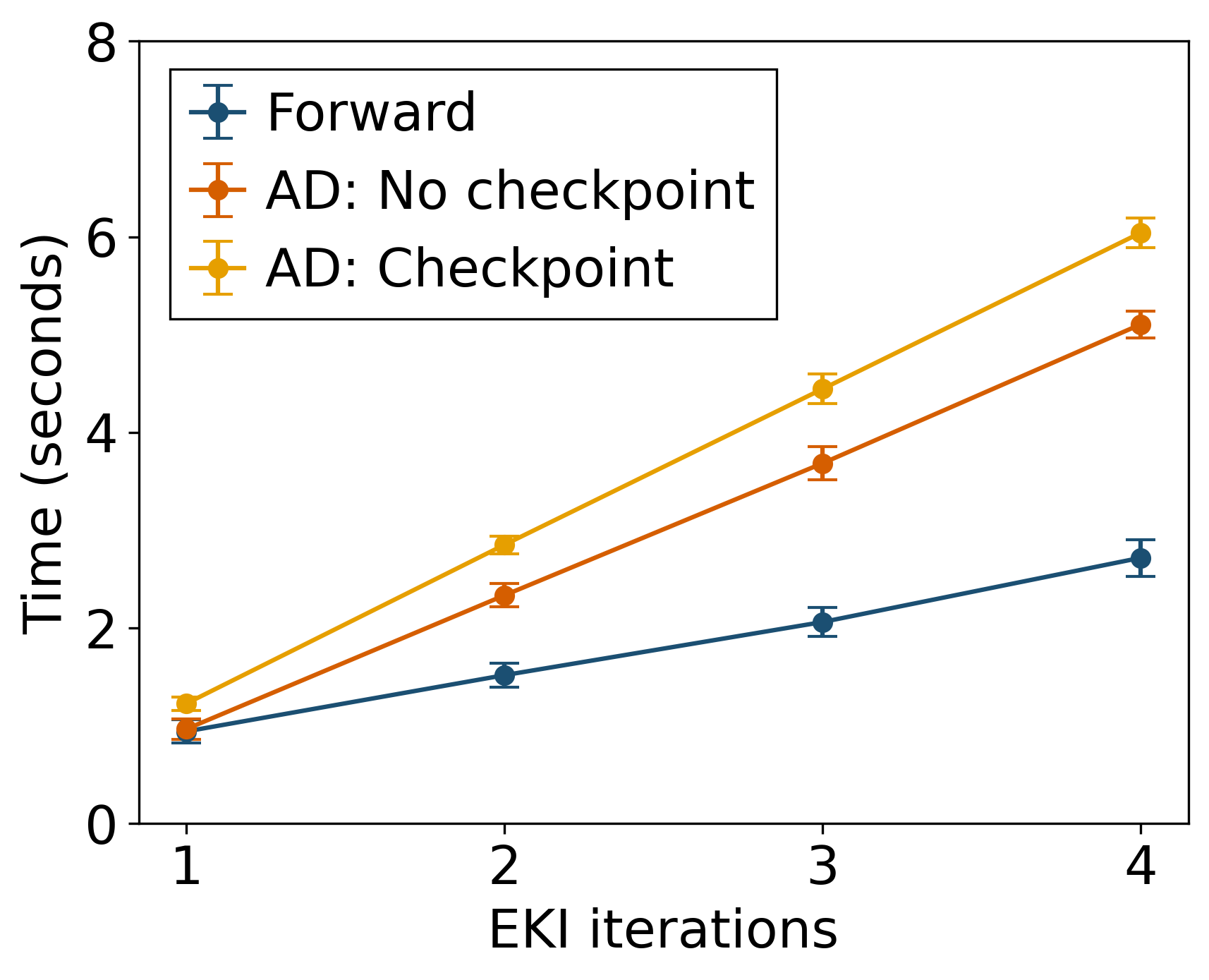}
    \caption{Time cost}
  \label{fig: Time cost step}
  \end{subfigure}
  \caption{Comparison of computation cost under different strategies across different numbers of EKI iteration. }
  \label{fig: cost - step}
\end{figure}

In this work, the AD-EKI introduced in Sections~\ref{sec:AD_EKI_KL} and~\ref{sec:AD_EKI_EIG} accounts for a critical component that enables the hybrid BED framework that can efficiently and robustly handle the model discrepancy. Here we conduct a comprehensive numerical study of the computational cost, including both peak memory usage and time consumption, for AD-EKI under varying ensemble sizes and different numbers of EKI iterations, as shown in Fig.~\ref{fig: cost - ensemble size} and Fig.~\ref{fig: cost - step}.

When the ensemble size is varied, the peak memory cost increases linearly with the ensemble size as shown in Fig.~\ref{fig: memory cost size}). This is because the number of evaluations for the forward model and the backpropagation both scale linearly with the ensemble size. The time cost also shows a linear increase roughly as presented in Fig.~\ref{fig: Time cost size}. For each ensemble member, the forward model evaluations are performed in parallel, meaning that the forward simulation time remains constant per iteration. The increasing time cost is therefore primarily attributed to matrix operations involved in EKI updates and the computation of the KL divergence, which overall scales linearly. It should be noted that the trend of time cost in Fig.~\ref{fig: Time cost size} would not hold true for problems whose forward model is very computationally expensive. In that case, the computational cost of the forward model is dominant and we should expect a roughly constant time cost with respect to the increased ensemble size.

On the other hand, if the number of EKI iterations is varied, the peak memory cost behaves differently depending on the differentiation strategy, with detailed comparison results summarized in Fig.~\ref{fig: memory cost step}. For standard forward EKI without automatic differentiation (AD), the memory cost is independent of the number of iterations with proper implementation, since no computational graph needs to be retained across iterations to facilitate AD. When the naive implementation of AD without checkpointing is applied, the memory cost increases linearly with the number of iterations, as the entire computational graph for all iterations must be stored. In contrast, when checkpointing is enabled and each EKI step is treated as a checkpoint, the memory cost remains constant after the first iteration. The time cost for forward EKI scales linearly with the number of iterations due to the iterative nature of the algorithm (Fig.~\ref{fig: Time cost step}). With AD enabled, the time cost increases, but remains within the same order of magnitude as forward EKI. Checkpointing further increases the overall time cost slightly due to recomputing certain quantities during backpropagation. Nevertheless, the time cost still maintains a linear relationship with the number of iterations.

\section{Discussion}
A key consideration is the robustness of the ensemble-based KL estimation, which relies on Gaussian assumptions and a finite ensemble size in practice. The ensemble-based KL divergence approximation in Eq.~\eqref{eq: eki updating} is indeed formally derived under Gaussian assumptions. However, numerous studies have demonstrated that this hybrid Monte Carlo/Kalman-gain approach remains remarkably robust even when the underlying posterior distribution is mildly non-Gaussian~\cite{evensen2003ensemble, roth2017ensemble, evensen2022data}. We acknowledge that performance can be degraded for strongly skewed or multimodal posteriors, yet the recent successes in applying ensemble Kalman methods to neural network parameter estimation~\cite{chen2019approximate, lopez2022training} suggest that much less complicated posteriors are often expected in real applications. Some research indicates that the posterior generated by EKI updates can become biased or eventually collapse into pure noise in later iterations~\cite{pensoneault2024efficient}, and our early stopping strategy effectively ensures the posterior remains informative and effective for the specific task of guiding experimental design.

Beyond the Gaussian assumption, challenges related to small ensemble sizes also warrant consideration. Insufficient ensemble sizes restrict the sample covariance to a low-dimensional subspace of the full parameter space, which restricts Kalman updates to that subspace and thus cannot correct errors orthogonal to it, resulting in biased estimates, underestimation of uncertainty, and potential divergence or slowed convergence of the ensemble Kalman methods. Given this potential issue of using ensemble Kalman methods, various techniques have been developed to mitigate challenges arising from small ensembles, including covariance inflation~\cite{anderson1999monte}, localization~\cite{gaspari1999construction, houtekamer1998data}, dropout~\cite{liu_dropout_2024}, and other covariance estimation methods~\cite{nino2015ensemble, ledoit2004well, vishny_high-dimensional_2024}. Most of those techniques can be straightforwardly integrated into our framework to further enhance its robustness against non-Gaussianity or small ensemble effects. As the core contribution of this paper is to propose a general framework for applying EKI to BED, we leave the detailed integration and evaluation of these specific enhancement techniques as a direction for future work.

One potential concern in BED is that the EIG can be highly sensitive to the choice of prior. More specifically, if the prior is misspecified, an EIG-based design may preferentially collect data that simply reinforce those incorrect beliefs, yielding little practical insight~\cite{dasgupta1991robust,muller2012measuring,go2022robust}. In our AD-EKI framework, this sensitivity is substantially reduced in both physical-parameter and network-parameter settings. For physical parameters, we begin with a broad range of uniform priors that are weakly informative and periodically recondition on an improved forward model, so that any bias in early posteriors is progressively mitigated as more accurate model corrections become available. More results can be found in~\ref{apx: sensitivity to prior}. For high-dimensional network parameters, ensemble Kalman methods feature a theoretical invariance, i.e., if the affine span of the initial ensemble covers the entire parameter space, then the posterior mean can, in principle, locate to any true parameter point, regardless of the prior’s high-density region. This implies that AD-EKI is theoretically insensitive to the specific shape of the network prior and depends only on subspace coverage. In practice, finite ensemble sampling introduces Monte Carlo errors, while AD-EKI allows resampling of the ensemble at each gradient update to mitigate these errors.

In terms of the scalability with respect to network size, i.e., the network parameter dimension \(d_{\theta}\), AD-EKI requires \(\mathcal{O}(J\,d_{\theta})\) memory, with \(\mathcal{O}(d_{\theta}^{2})\) for standard EKI and \(\mathcal{O}(J\,d_{\theta})\) for dropout-EKI. In terms of convergence, the ensemble-based gradient estimate in AD-EKI retains an \(O(J^{-1/2})\) error bound, which is the same as EKI since the KL divergence between Gaussians is Lipschitz continuous in both mean and covariance. Rigorous proof can be found in the related work AD-EnKF~\cite{chen_autodifferentiable_2022}. Though this is explicitly independent of \(d_{\theta}\), typically standard EKI requires a similar amount of $J$ as $d_\theta$ to achieve this convergence rate. Dropout, or other localized covariance estimation techniques, may relax this requirement. Stability of AD-EKI stems from using derivative-free covariance updates in the inner loop of this bilevel optimization problem and from limiting the differentiation graph depth to a modest number of ensemble iterations. When the network dimension is huge such that drawing enough ensemble members is infeasible, additional techniques such as covariance inflation, dropout-based ensemble diversification, or other methods that break the subspace constraint may be applied to improve the robustness and stability of ensemble Kalman methods, thus facilitating the efficient use AD-EKI in real-world large-scale problems.

\section{Conclusion}\label{Conclusion}
In this work, we propose a hybrid Bayesian experimental design framework, enabled by auto-differentiable ensemble Kalman inversion (AD-EKI), to iteratively identify optimal designs for determining unknown physical parameters and model discrepancy. We validate the performance of the proposed method with a classical numerical example for BED governed by a convection-diffusion equation. More specifically, we first study the example with a low-dimensional parametric model error, for which full Bayesian inference of the joint unknown parameters is still feasible. The results confirm that the proposed method can effectively calibrate the parametric model error based on the optimal designs identified by the hybrid framework with AD-EKI, which consequently leads to satisfactory performance of inferring the physical parameters. The proposed method is then studied in the same numerical example with a high-dimensional structural model error, for which standard BED methods via full Bayesian inference can be infeasible. Instead of empirically choosing a large amount of data to calibrate the model discrepancy offline, we demonstrate that the hybrid framework with AD-EKI identifies informative data online to calibrate high-dimensional model discrepancy and ensure good performance of inferring the physical parameters. These findings confirm that the proposed method provides an efficient and robust approach for Bayesian experimental design with model discrepancy. It is worth noting that AD-EKI can also potentially benefit the development of efficient and scalable differentiable programming frameworks for many other problems that involve bilevel (nested) optimization in various areas such as machine learning (e.g., hyperparameter optimization, meta-learning), structure optimization (e.g., formulated as mathematical programming problems with equilibrium constraint), economic problems (e.g., toll
setting), and defense applications (e.g., strategic offensive and defensive systems), by handling the inner (lower) optimization via AD-EKI.

\section*{Acknowledgments}
H.Y., X.D., and J.W. are supported by the University of Wisconsin-Madison, Office of the Vice Chancellor for Research and Graduate Education with funding from the Wisconsin Alumni Research Foundation.

\section*{Data Availability}
The data that support the findings of this study are available from the corresponding author upon reasonable request.
  
\bibliographystyle{unsrt}
\bibliography{references}

\clearpage
\appendix
 \section{Details of derivation for the gradient of ensemble-based KL divergence}\label{apd:derivation}
To show the derivation of the detailed form of Eq.~\eqref{eq: EKI gradient 1}, we start with the KL divergence between two multivariate Gaussians:
\[
D_{\mathrm{KL}}(p(\boldsymbol{\theta}|\mathbf{y}) || p(\boldsymbol{\theta})  )
= \frac{1}{2} \left[
    \operatorname{tr}(\Sigma_0^{-1} \Sigma_K)
    + (\bar{\bm\theta}_0 - \bar{\bm\theta}_K)^\top \Sigma_0^{-1} (\bar{\bm\theta}_0 - \bar{\bm\theta}_K)
    - d
    + \log \frac{\det \Sigma_0}{\det \Sigma_K}
\right]
\]

Define the quadratic term:
\[
T_3 := (\bar{\bm\theta}_0 - \bar{\bm\theta}_K)^\top \Sigma_0^{-1} (\bar{\bm\theta}_0 - \bar{\bm\theta}_K),
\]
whose gradient with respect to \(\bar{\bm\theta}_K\) is:
\[
\frac{\partial T_3}{\partial \bar{\bm\theta}_K}
= -2 \Sigma_0^{-1} (\bar{\bm\theta}_0 - \bar{\bm\theta}_K).
\]

Thus, the partial derivative of the KL divergence with respect to \(\bar{\bm\theta}_K\) is:
\[
\frac{\partial D_{\mathrm{KL}}}{\partial \bar{\bm\theta}_K}
= -\Sigma_0^{-1} (\bar{\bm\theta}_0 - \bar{\bm\theta}_K)
= \Sigma_0^{-1} (\bar{\bm\theta}_K - \bar{\bm\theta}_0),
\]
which accounts for the detailed form of the first term at the right-hand side in Eq.~\eqref{eq: EKI gradient 1}. To show how the second term is derived, we introduce two identities:

\textbf{(Identity \#1)} \(\displaystyle \mathrm d\,\tr(A X) = \tr\bigl(A\,\mathrm dX\bigr)\)  
   \(\displaystyle \Longrightarrow \frac{\partial}{\partial X}\tr(A X)=A^\top.\)

\textbf{(Identity \#2)} \(\displaystyle \mathrm d\,\ln\det X = \tr\bigl(X^{-1}\,\mathrm dX\bigr)\)  
   \(\displaystyle \Longrightarrow \frac{\partial}{\partial X}\ln\det X=(X^{-1})^\top.\)  
   If \(X\) is symmetric, \((X^{-1})^\top=X^{-1}\).  
   The trace ensures the differential is a scalar.

We first define
\[
T_1 := \tr(\Sigma_0^{-1}\Sigma_K),
\quad
T_2 := \ln\det\Sigma_K.
\]

Using Identity \#1, we can have
\[
\frac{\partial T_1}{\partial \Sigma_K}
= (\Sigma_0^{-1})^\top
= \Sigma_0^{-1},
\quad
\mathrm d\,T_1 = \tr\bigl(\Sigma_0^{-1}\,\mathrm d\Sigma_K\bigr).
\]

Using Identity \#2, we can have:
\[
\frac{\partial T_2}{\partial \Sigma_K}
= (\Sigma_K^{-1})^\top
= \Sigma_K^{-1},
\quad
\mathrm d\,T_2 = \tr\bigl(\Sigma_K^{-1}\,\mathrm d\Sigma_K\bigr).
\]

Therefore, we have the following derived results
\[
\frac{\partial D_{\mathrm{KL}}}{\partial \Sigma_K}
= \tfrac12\bigl(
  \Sigma_0^{-1} - \Sigma_K^{-1}
\bigr),
\]
which accounts for the detailed form of the second term at the right-hand side in Eq.~\eqref{eq: EKI gradient 1}.

\section{Sensitivity test to prior of physical parameters}\label{apx: sensitivity to prior}

As mentioned in the Discussion, prior sensitivity is one concern in BED~\cite{dasgupta1991robust,muller2012measuring,go2022robust}. In this section, we provide results with different priors of physical parameters to validate our argument about the insensitivity of our framework to priors.

In the numerical tests in Sections~\ref{sec:Correct parametric error} and~\ref{sec:Correct functional error}, we have already demonstrated that our method works well with uniform priors.

\begin{figure}[H]
    \centering
    \includegraphics[width=0.8\linewidth]{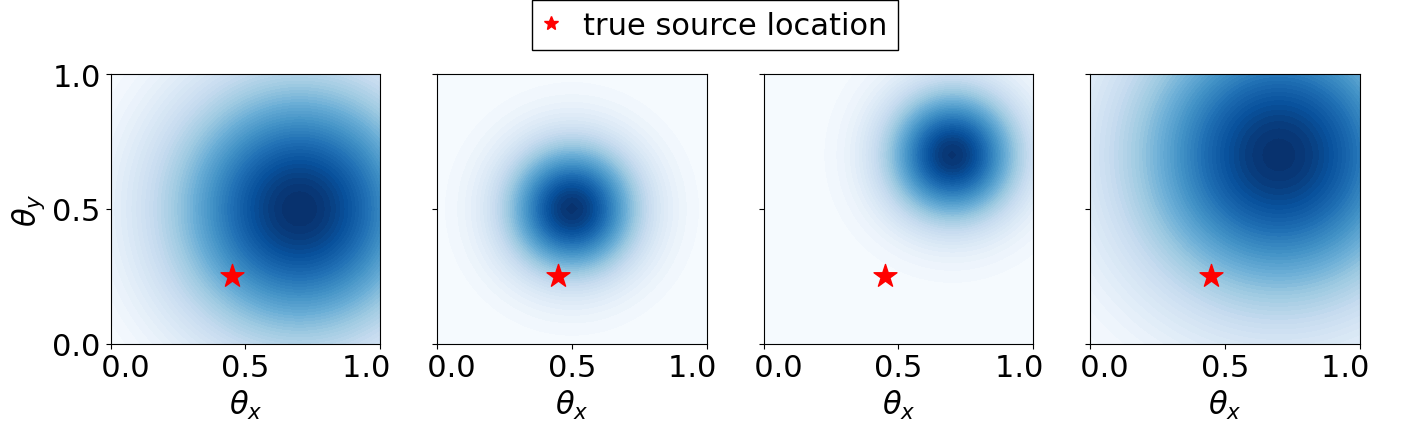}
    \caption{Different 2D Gaussian priors for the physical parameters. From left to right: (1) flat and broad prior, (2) peaked prior centered near the true value, (3) peaked prior centered far from the true value, and (4) initially peaked and biased prior flattened into a broader distribution.}
    \label{fig: priors}
\end{figure}

\begin{figure}[H]
    \centering
    \includegraphics[width=0.8\linewidth]{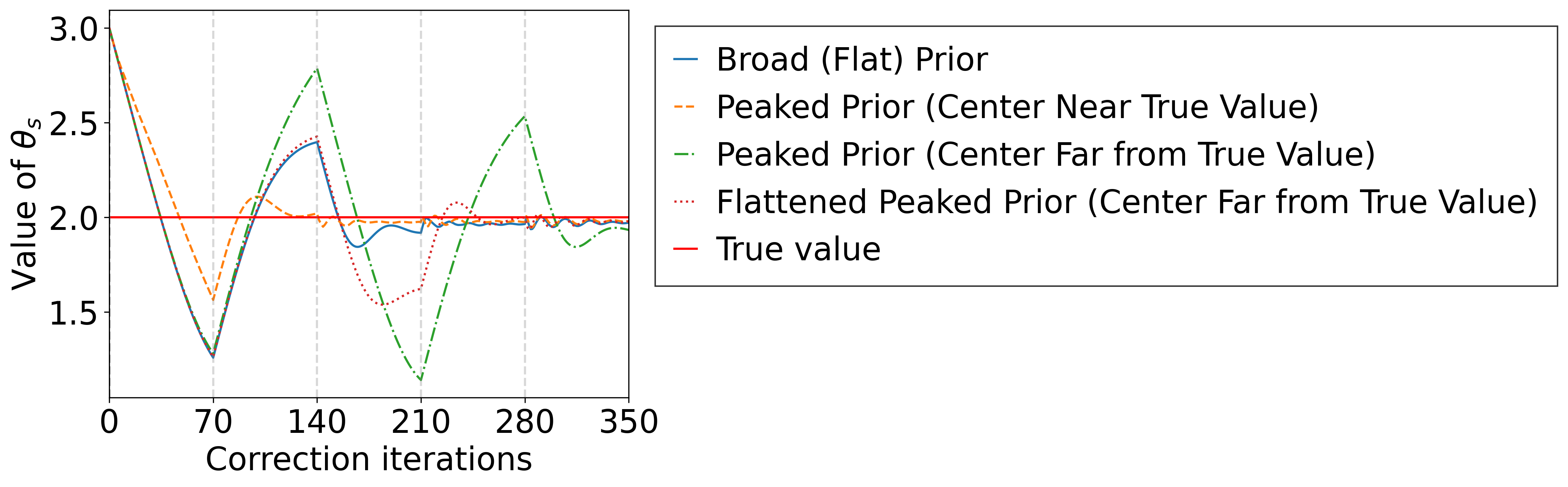}
    \caption{Error parameter correction trajectory with different priors.}
    \label{fig: sensitivity theta s}
\end{figure}

\begin{figure}[H]
    \centering
    \includegraphics[width=\linewidth]{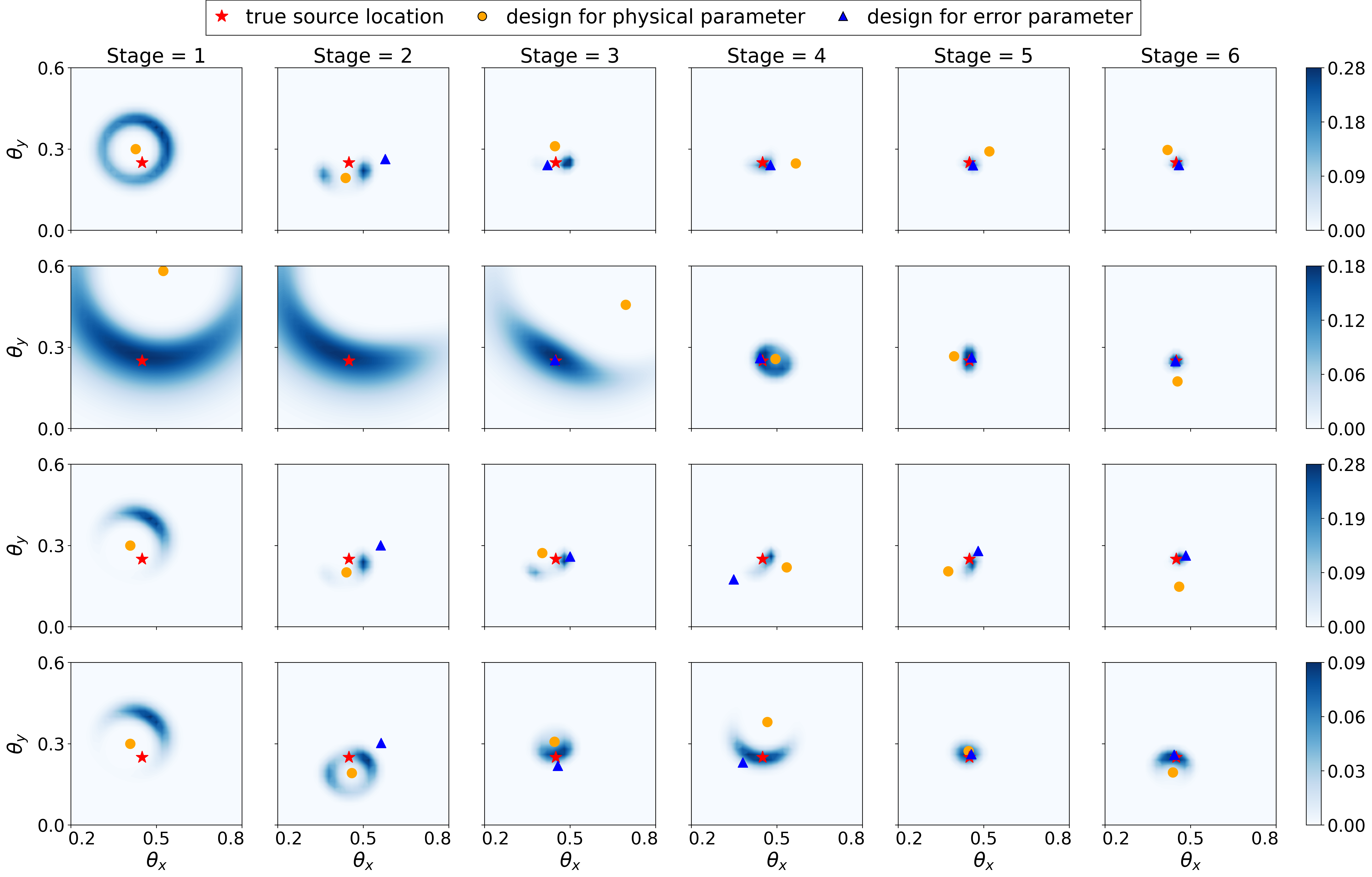}
    \caption{Posterior results of the AD-EKI framework under different physical priors. From top to bottom: (1) flat and broad prior, (2) peaked prior centered near the true value, (3) peaked prior centered far from the true value, and (4) peaked and biased prior flattened into a broader distribution.}
    \label{fig: sensitivity posterior}
\end{figure}

We further consider the misspecified, which is biased, priors in the parametric case. In particular, we consider four types of misspecified 2D Gaussian priors: (1) flat and broad, (2) peaked and centered near the true value, (3) peaked and centered far from the true value, and (4) initially peaked and biased, but subsequently flattened into a near-uniform distribution, as illustrated in Fig.~\ref{fig: priors}. Specifically, we obtain case (4) by raising each density value of the case (3) prior to the $1/5$ power and then renormalizing so that the transformed distribution sums to one. The error correction trajectories (Fig.~\ref {fig: sensitivity theta s}) and posterior distributions (Fig.~\ref{fig: sensitivity posterior}) in cases (1) and (2) demonstrate that our method remains robust under misspecified but moderately inaccurate priors. These results indicate a low sensitivity to priors. Although the convergence rate is slow under strongly biased priors (case 3), flattening such priors into a broader distribution (case 4) restores the effectiveness of the AD-EKI framework. Flattening the prior essentially places less weight on the prior and more on the data, effectively transforming a biased prior into a near-uniform one, under which our method has already been shown to be effective. 

These findings reinforce the practical robustness of our method under a wide range of prior assumptions, justifying its applicability in scenarios with uncertain or biased prior knowledge.

\end{document}